\documentclass[10pt,twocolumn,letterpaper]{article}

\usepackage{cvpr}
\usepackage[accsupp]{axessibility} 
\usepackage{times}
\usepackage{epsfig}
\usepackage{graphicx}
\usepackage{amsmath}
\usepackage{amssymb}
\usepackage{booktabs}

\usepackage{pifont}

\usepackage{tabularx}
\usepackage{soul}

\def\wacvPaperID{0098} 

\cvprfinalcopy 

\usepackage[pagebackref=true,breaklinks=true,colorlinks,bookmarks=false]{hyperref}
\hypersetup{
    colorlinks=true,
    linkcolor=black,
    filecolor=magenta,      
    urlcolor=cyan,
    pdftitle={Fine-grained Activities of People Worldwide},
    pdfpagemode=FullScreen,
}

\usepackage{verbatim}
\usepackage{times}
\usepackage{epsfig}
\usepackage{graphicx}
\usepackage{amsmath, amsthm, amssymb}
\usepackage{mdwlist}
\usepackage{mdframed}
\usepackage[ruled,vlined]{algorithm2e}
\usepackage{authblk}
\usepackage{etoolbox}

\newenvironment{squishlist}
 {\begin{list}{$\bullet$}
  { \setlength{\itemsep}{0pt}
     \setlength{\parsep}{3pt}
     \setlength{\topsep}{3pt}
     \setlength{\partopsep}{0pt}
     \setlength{\leftmargin}{1.5em}
     \setlength{\labelwidth}{1em}
     \setlength{\labelsep}{0.5em} } }
{\end{list}}

\newtoggle{arXiv}
\toggletrue{arXiv}   
\newtoggle{doubleblind}
\togglefalse{doubleblind}   

\newcommand{\figwidth}{6.85in}
\newcommand{\fighalfwidth}{3.4in}
\newcommand{\figthirdwidth}{2.25in}

\begin{document}

\title{Fine-grained Activities of People Worldwide}

\author[1]{Jeffrey Byrne}
\author[2]{Greg Casta\~{n}\'{o}n}
\author[2]{Zhongheng Li}
\author[2]{Gil Ettinger}
\affil[1]{Visym Labs, Cambridge MA, USA}
\affil[2]{Systems \& Technology Research, Woburn MA, USA}


\maketitle

\thispagestyle{empty}

\begin{abstract}
Every day, humans perform many closely related activities that involve subtle discriminative motions, such as putting on a shirt vs. putting on a jacket, or shaking hands vs. giving a high five.  Activity recognition by ethical visual AI could provide insights into our patterns of daily life, however existing activity recognition datasets do not capture the massive diversity of these human activities around the world.  To address this limitation, we introduce Collector, a free mobile app to record video while simultaneously annotating objects and activities of consented subjects.  This new data collection platform was used to curate the Consented Activities of People (CAP) dataset, the first large-scale, fine-grained activity dataset of people worldwide. The CAP dataset contains 1.45M video clips of 512 fine grained activity labels of daily life, collected by 780 subjects in 33 countries.  We provide activity classification and activity detection benchmarks for this dataset, and analyze baseline results to gain insight into how people around with world perform common activities. 
The dataset, benchmarks, evaluation tools, public leaderboards and mobile apps are available for use at
\iftoggle{doubleblind} {\href{https://}{REDACTED}}{\href{https://visym.github.io/cap}{https://visym.github.io/cap}}.
\end{abstract}

\section{Introduction}
\label{s:introduction}

Large scale activity recognition has made remarkable progress driven by the curation of large scale labeled video datasets \cite{Karpathy2014,Soomro2012, Heilbron2015,abu2016youtube, Alayrac2016, Xu2016, Krishna2017, Carreira2017,Fouhey2018, Jiang2018, Monfort2020, jia2020lemma, liu2017pku, Kong_2019_ICCV}.  Evaluation tasks in these datasets include activity classification, activity and object detection and localization, action prediction, episodic memory for object instance retrieval, object interactions with hands/tools/people, speaker prediction and scene diarization in long duration videos.

However, performance on these important tasks remains limited by the scale, quality and applicability of data.  While there are many large-scale video datasets for pretraining activity recognition such as 
Kinetics \cite{kay2017kinetics},
AVA \cite{gu2018ava}, 
Moments in Time \cite{Monfort2020}\cite{9609554},
ActivityNet \cite{caba2015activitynet}, YouTube-8M \cite{abu2016youtube}, HVU \cite{hvu} and 
IG65M \cite{Ghadiyaram19}, these datasets are all scraped from social media platforms such as YouTube or Instagram.  These datasets are easy to collect, but suffer from terms of service restrictions, non-consented subjects and link rot, making reproducible research difficult.  Furthermore, the labels in these datasets sparsely sample fine-grained activities, and instead represent activities that are interesting enough for social media.  Recent dataset collections efforts have transitioned to actors performing scripted \cite{jia2020lemma}\cite{Rai21} or unscripted \cite{Ego4D2021} activities to introduce more diversity of the activities that we all perform every day, however these datasets have limited scale for supervised training. 

\begin{figure}[t!]
\centering
\includegraphics[width=3.25in]{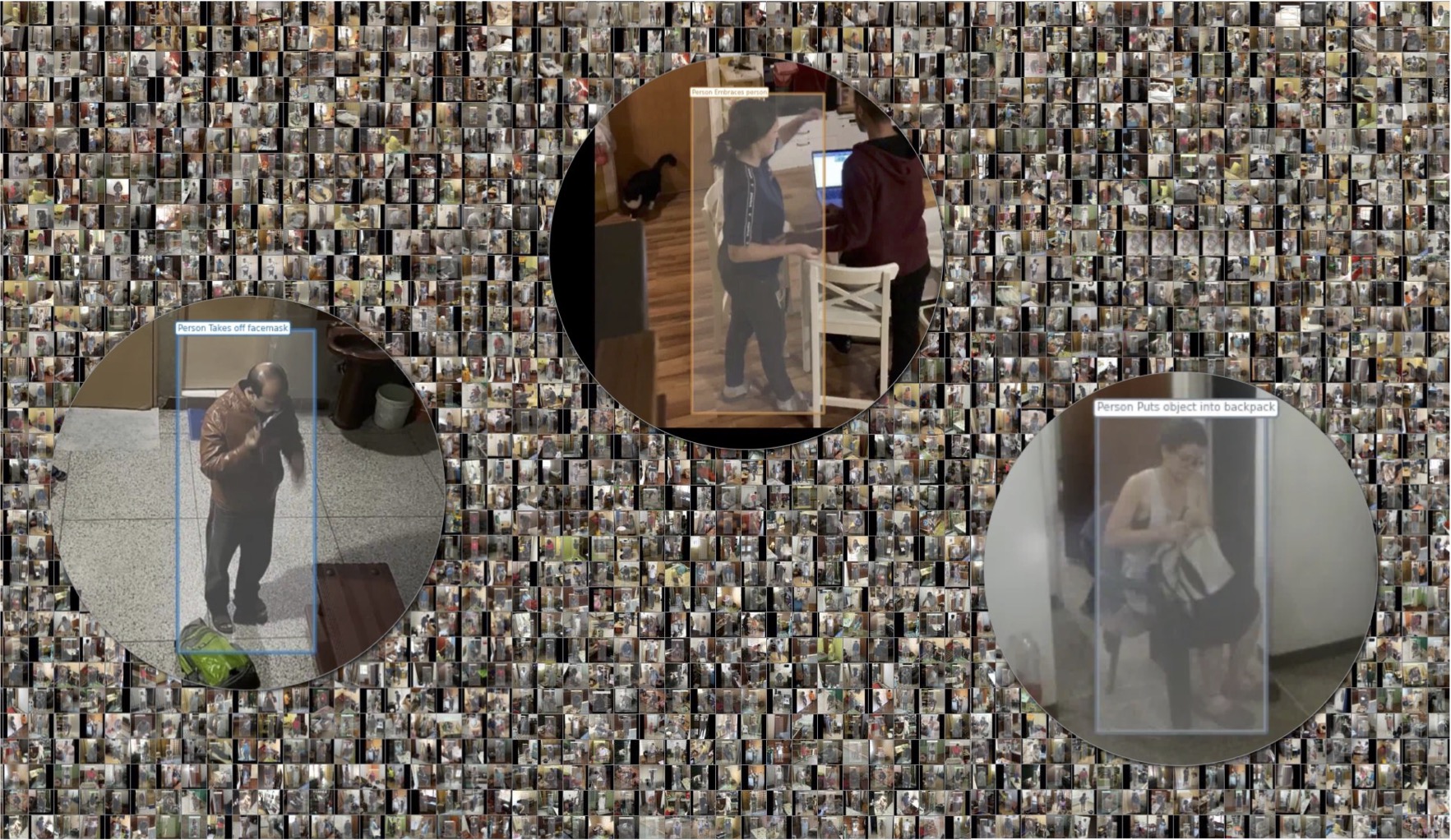}
\caption{The Diversity of Human Activities.  Humans perform 
\iftoggle{doubleblind} {a wide variety}{\href{https://youtu.be/HjNa7_T-Xkc}{a wide variety}} 
of 
\iftoggle{doubleblind} {closely related activities}{\href{https://youtu.be/KvULvetOV8c}{closely related activities}} 
that involve subtle motions performed alone, while interacting with objects or with other people.  The
\iftoggle{doubleblind} {CAP dataset}{\href{https://visym.com/cap}{CAP dataset}}  
was designed to explore the representation of these fine grained activities of daily life around the world.}
\label{f:cap_explorer}
\end{figure}

\begin{figure*}
\centering
\includegraphics[width=\figwidth]{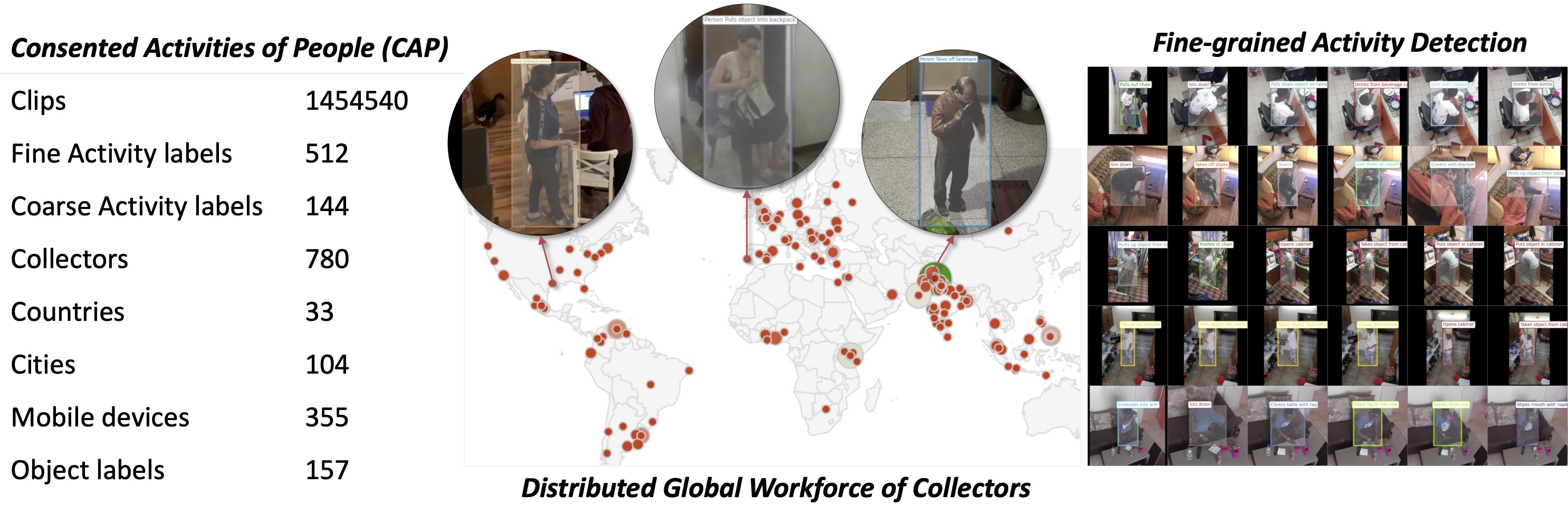}
\caption{The Consented Activities of People Dataset, 
collected on-demand from consented subjects, recorded worldwide from third-person viewpoints, of fine-grained activities of daily life and submitted from handheld and rigid mobile devices.  Available at 
\iftoggle{doubleblind} {\href{https://}{{\bf REDACTED}}}{\href{https://visym.com/cap}{visym.com/cap}}.}
\label{f:cap_statistics}
\end{figure*}

In this paper, we introduce the Consented Activities of People (CAP) dataset,  a fine grained dataset of activities of daily life for visual AI research.  Humans perform a wide variety of closely related activities that involve simple yet subtle motions that we perform alone, interacting with objects or interacting with other people.  For example, figures \ref{f:cap_explorer} and \ref{f:cap_statistics} show examples of activities that we may perform every day: putting on a face mask, putting an object into a backpack or hugging another person.  The CAP dataset was designed to explore the representation and recognition of fine grained activities of daily life, using open data collected on-demand from consented subjects and recorded worldwide from third-person viewpoints.  Specifically, the CAP dataset contains:

\smallskip
\begin{squishlist}
\item {\textbf{Common activities}} that we all perform each day, such as dressing or grooming that are not typically captured on video because they are rarely performed in front of a camera or are too boring to share. 
\item {\textbf{Fine activities}} that are closely related and may be easily confused, such as putting on socks vs. putting on shoes or talking on a phone vs. smoothing your hair.
\item {\textbf{Diverse activities}} that are different ways of performing the same activity in the wild, such as activities viewed from behind or interacting with different objects.
\end{squishlist}
\smallskip

In order to collect a large-scale visual dataset of the diversity of human activities, we introduce the {\em Collector platform}.  
Collector is a global platform for collecting consented datasets of people for visual AI applications. Collector is able to record, annotate and verify video datasets, collected with geographically diversity of people around the world.

The primary contributions of this work are:

\smallskip
\begin{squishlist}
    \item {\bf Collector platform}.  Section \ref{s:collector} describes the new platform developed to collect ethical datasets of people.  This platform can be used by the research community to collect new on-demand visual datasets as easily as recording a video.   
    \item {\bf Consented Activities of People (CAP) dataset}.  Section \ref{s:cap} describes the collected dataset of fine-grained activities of consented people worldwide.  The dataset contains annotated videos of fine-grained activities with bounding box tracks and temporal localization. 
    \item {\bf Benchmark suite}.  Section \ref{s:benchmark} describes the open benchmarks and baselines on this dataset, along with results and analysis in Section \ref{s:results}.  
\end{squishlist}

\section{Related Work}

\begin{table*}[]
\small
\begin{tabularx}{\textwidth}{llllllll}
\hline
Dataset & Year & Domain & Fine Classes & Coarse Classes & Annotation & Clips & Mean Clips/Class\\ \hline
ActivityNet (v3) \cite{caba2015activitynet} & 2016 & Social & 200 & 73 & T & 23.1K & 137\\
Charades \cite{Sigurdsson2016}\cite{Sigurdsson2018ActorAO} & 2016 & Exo$\vert$Ego & - & 157 & T$\vert$N & 68.5K & -\\
Something-Som... \cite{Goyal2017} & 2017 & Ego & - & 174 & T & 220.8K & 600 \\
PKU-MMD \cite{liu2017pku} & 2017 & Exo$\vert$M & - & 51 & T$\vert$J & 21.5K &  - \\
Kinetics-700 \cite{kay2017kinetics}  & 2017 & Social & - & 700 & T & 650K  & 700      \\
Youtube-8M (Seg) \cite{abu2016youtube} & 2018 & Social  & - & 1000 & T & 237K & 150 \\
EPIC-Kitchens \cite{Damen2018EPICKITCHENS} & 2018 & Ego & 97 & 13 & T$\vert$N$\vert$C & 90K & -\\
HACS (clips) \cite{zhao2019hacs}  & 2019 & Social  & - & 200 & T & 1.55M & 1100  \\
MMAct \cite{Kong_2019_ICCV} & 2019 & Exo$\vert$Ego$\vert$M & - & 37 & S$\vert$T$\vert$J & 40K & -\\
LEMMA \cite{jia2020lemma}& 2020 & Exo$\vert$Ego$\vert$M  & 863 & 24 & S$\vert$T & 11.8K & - \\
AVA-Kinetics \cite{AVA-Kinetics-2020} & 2020 & Social & - & 60  & S$\vert$T & 230K & 235 \\ 
HVU (Actions) \cite{hvu} & 2020 & Social & - & 739 & T & 479.5K & 2112 \\
Moments in Time \cite{9609554} & 2020 & Social & - & 292 & T & 2.01M & 6432 \\
MEVA \cite{Corona2021WACV}& 2021 & Exo & - & 37 & S$\vert$T$\vert$E$\vert$C & 35K & - \\
%
HOMAGE \cite{Rai21} & 2021 & Exo$\vert$Ego$\vert$M  & 453 & 70 & S$\vert$T & 24.6K &  - \\
Ego4D (MQ) \cite{Ego4D2021}  & 2021 & Ego & - & 110 & S$\vert$T$\vert$N$\vert$G$\vert$C & 22.2K & -\\ 
\hline
CAP  & 2022 & Exo & 512 & 144 & S$\vert$T$\vert$N$\vert$G$\vert$C & 1.45M & 2880 (4501, top-250)  
\end{tabularx}
\smallskip
\caption{CAP dataset comparison.  Domains are egocentric (ego) from a first person viewpoint such as a head or body mounted camera, exocentric (exo) from a third-person viewpoint such as from a wall or building mounted camera, (social) videos scraped from online social media sources and (M)ulti-modal domains such as RGB-D, NIR, multiple viewpoints or additional non-visual sensors.  Annotation ground truth considers combinations of: (T)emporal activity labels for start and end times, (S)patial object labels of bounding boxes around actors or interacted objects, (E)xtrinsic camera poses with calibrated relative position and orientation, (G)eographic locations for each video, (J)oint keypoints of human pose skeletons, (N)atural language captions or narrations and (C)onsented subjects for ethical video recording.  } 
\label{t:comparison}
\end{table*}

The evolution of video datasets has progressed from a small number of classes and actors in trimmed videos \cite{Schuldt2004a,Gorelick07} to large-scale web video on social media \cite{Karpathy2014,Soomro2012, Heilbron2015,abu2016youtube, Alayrac2016, Xu2016, Krishna2017, Carreira2017,Fouhey2018, Jiang2018, Monfort2020}.  Keyword-based search from  YouTube or Instagram enabled weak labeling of videos with minimal curation, creating datasets that recorded a large set of people doing a small set of activities.  The diversity and volume of video available on social media lead to massive datasets for pretraining.  Recently, efforts  have bootstrapped classifiers to improve the scalability of their annotation and collection efforts from noisy web video \cite{zhao2019hacs}.  Furthermore, approaches have attempted to directly mitigate the geographic biases of web video by scraping from local versions of websites \cite{aviddataset}.

These large datasets are easy to curate, but the contents have limited diversity, as the joint combination of viewpoints (e.g. exocentric, egocentric) and activity labels (e.g. dressing, eating) that are common in real scenes are not as common on social media.  Centralized collection of actors \cite{Corona2021WACV, Rai21,ji2020action}, as well as crowdsourced approaches \cite{Sigurdsson2016}\cite{Castanon19}\cite{1706.04261}\cite{Rai21}\cite{ji2020action} have been used to generate datasets of labels and perspectives not densely sampled in social video, but are limited in the diversity and scale of training data.  This style of dataset collection has specialized further into diagnostic datasets \cite{1612.06890}\cite{CLEVRER}\cite{1910.04744}\cite{objectnet}\cite{2007.04954} that attempt to answer a specific question about performance bias, as well as fine-grained datasets which attempt to densely sample the space of actions in a specific domain \cite{2105.11107, moca2020, 2105.07404, 2105.07404, 1804.03247, 6247801, 1905.04430, Jones2020}.

Table \ref{t:comparison} shows a quantitative comparison of these related datasets.  This comparison table focuses on egocentric, exocentric and social datasets for activity classification and detection tasks, comparing the number of classes, clips and mean clips per class.  
This shows that our Consented Activities of People (CAP) dataset is the largest consented activity dataset collected to date as measured by mean number of training clips per class.

\begin{figure*}[t!]
    \centering
    \iftoggle{doubleblind} {\includegraphics[width=\figwidth]{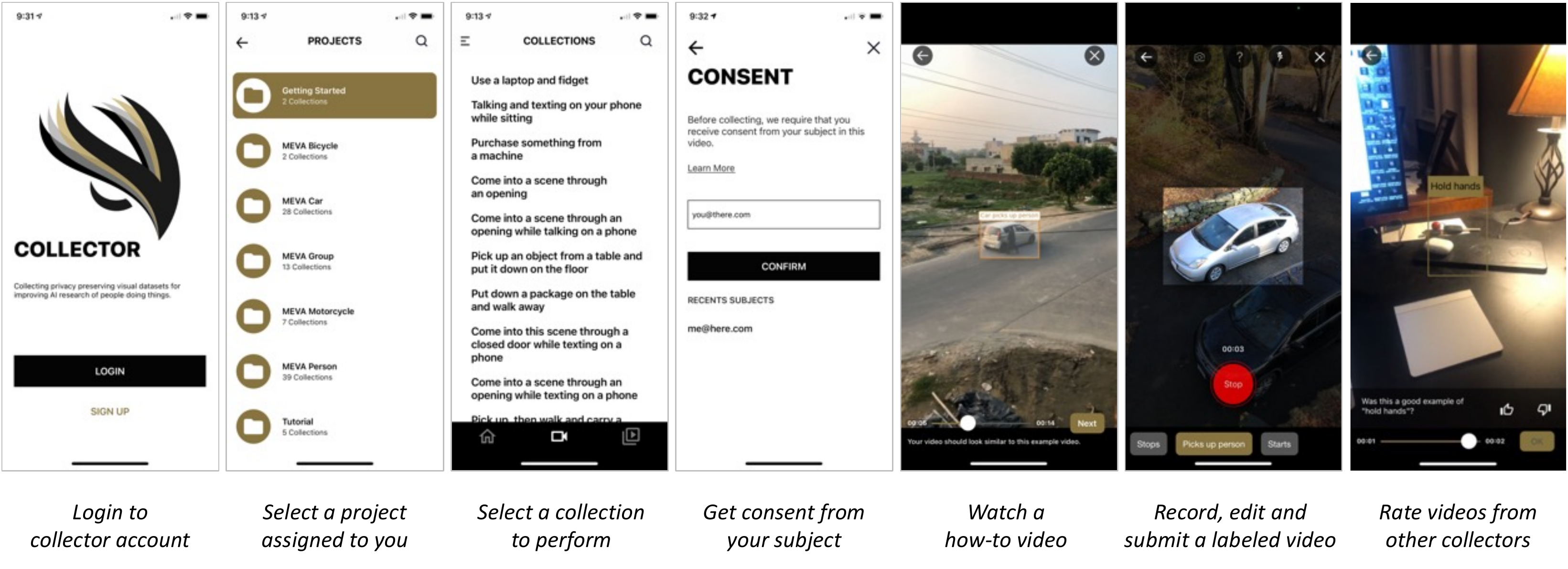}}
    {\includegraphics[width=\figwidth]{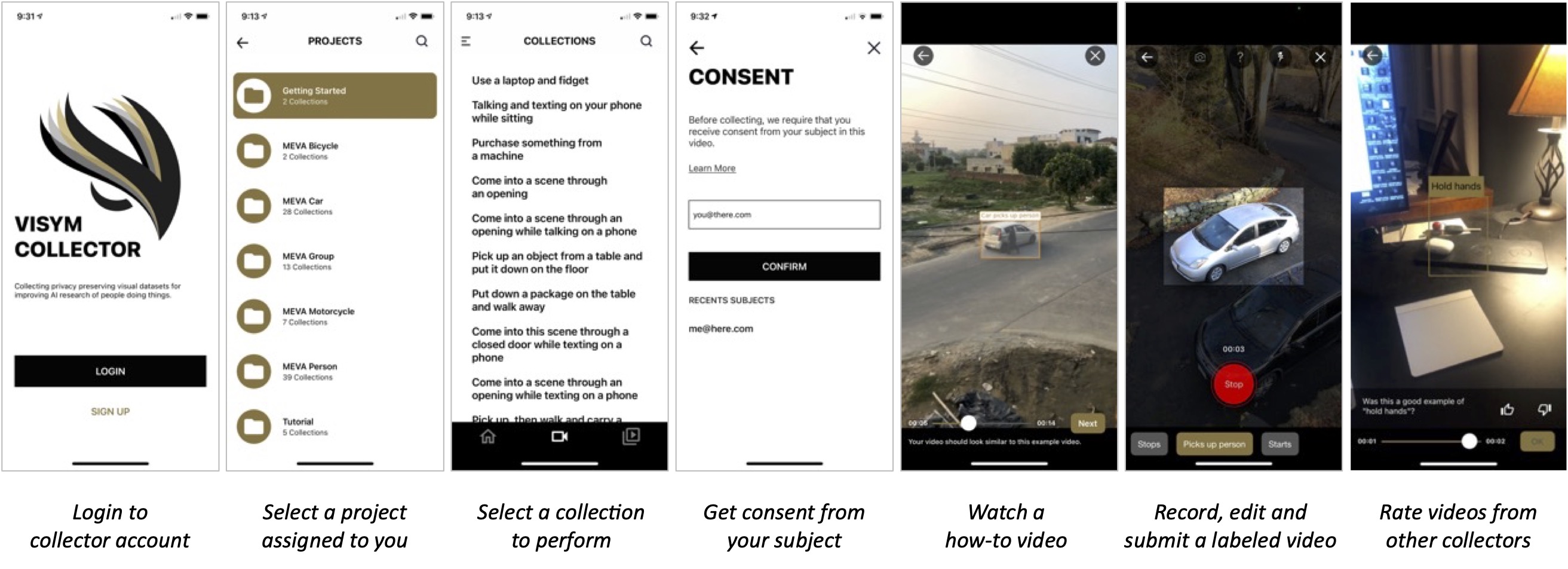}
    }
    \caption{The 
    \iftoggle{doubleblind} {Collector platform}{\href{https://visym.com/collector}{Collector platform}}
    curates visual datasets of people by enabling thousands of collectors worldwide to record and submit videos using a mobile app. This workflow shows the mobile interface for collecting on-demand video datasets of people.}
    \label{f:collector_workflow_phase3}
\end{figure*}

\section{Collector Platform}
\label{s:collector}

Collector is a new platform for visual dataset curation that was designed to address the limits of current collection strategies.
The traditional approach to construction of visual dataset of people is to: (i) Set up camera networks to record videos and imagery, (ii) Gather a set of subjects who have consented to have their personally identifiable information (PII) recorded and shared for an authorized purpose and duration, (iii)  Record videos of these IRB approved consented subjects only, and no one else, (iv) Send videos to an annotation team to manually search videos for ground truth labels, (v) Send the annotations to a verification team to enforce quality.  This approach is slow end expensive.  

There is a need for a new dataset collection approach that is {\em on-demand, worldwide and cost-efficient}.  On-demand approaches enable an agile, adaptive collection of  instances that are engineered to introduce diversity of labels or attributes such as pose, illumination or object interaction and mitigate biases.  Furthermore, access to data sources from many countries and cultures avoids an imbalance of data from a specific region of the world and its implicit biases.  Finally, the approach needs to be relatively cost-efficient to collect large-scale training data.  

Collector is a global platform for collecting large scale consented video datasets of people for visual AI applications. Collector is able to record, annotate and verify custom video datasets of rarely occurring activities for training visual AI systems. 
The Collector platform provides:

\smallskip
\begin{squishlist}
\item On-demand collection of rarely occurring activities from thousands of collectors worldwide.
\item Simultaneous video recording, annotation and verification into a single unified platform.
\item Touchscreen UI for live annotation of bounding boxes, activity clips and object categories.
\item Specification of required collection attributes such as pose, illumination, location or object interactions. 
\item IRB approved informed consent for ethical dataset construction with in-app face anonymization.
\end{squishlist}
\smallskip

Figure \ref{f:collector_workflow_phase3} shows an overview of the collector workflow.  Collectors are invited onto the platform, and they download the collector mobile app to their device.  Collectors are presented \emph{collections}  which are video collection tasks grouped by required objects (e.g. a car, another person) or locations (e.g. parking lot, dining room). Each collection specifies the requirements of the submitted video, which include required activities, objects, location, illumination conditions, actor pose and camera viewpoint.  Once a collector chooses a collection to record, they get consent from their subject, including a video recording to ensure that the person consenting is the person being recorded.  Next, the collector watches an example video which shows a gold standard exemplar of the collection.  We use visual exemplars to bypass language issues and communicate an idea of what the collection should look like.  Finally, the collector records and annotates the video live using touch gestures on their device, optionally corrects errors using an in-app annotation editor and submits the annotated collection for review. 

The Collector mobile app has been downloaded by thousands of freelance collectors worldwide, and is freely available in the iOS and Android app stores.  Appendix \ref{a:collector} provides more information on mobile app for recording and annotation (\S\ref{a:mobile_app}), campaign dashboard for global coordination (\S\ref{a:campaign_dashboard}) and human review for annotation quality (\S\ref{a:human_review}).  

\begin{figure*}[t!]
\centering
\iftoggle{doubleblind} {\includegraphics[width=\figwidth]{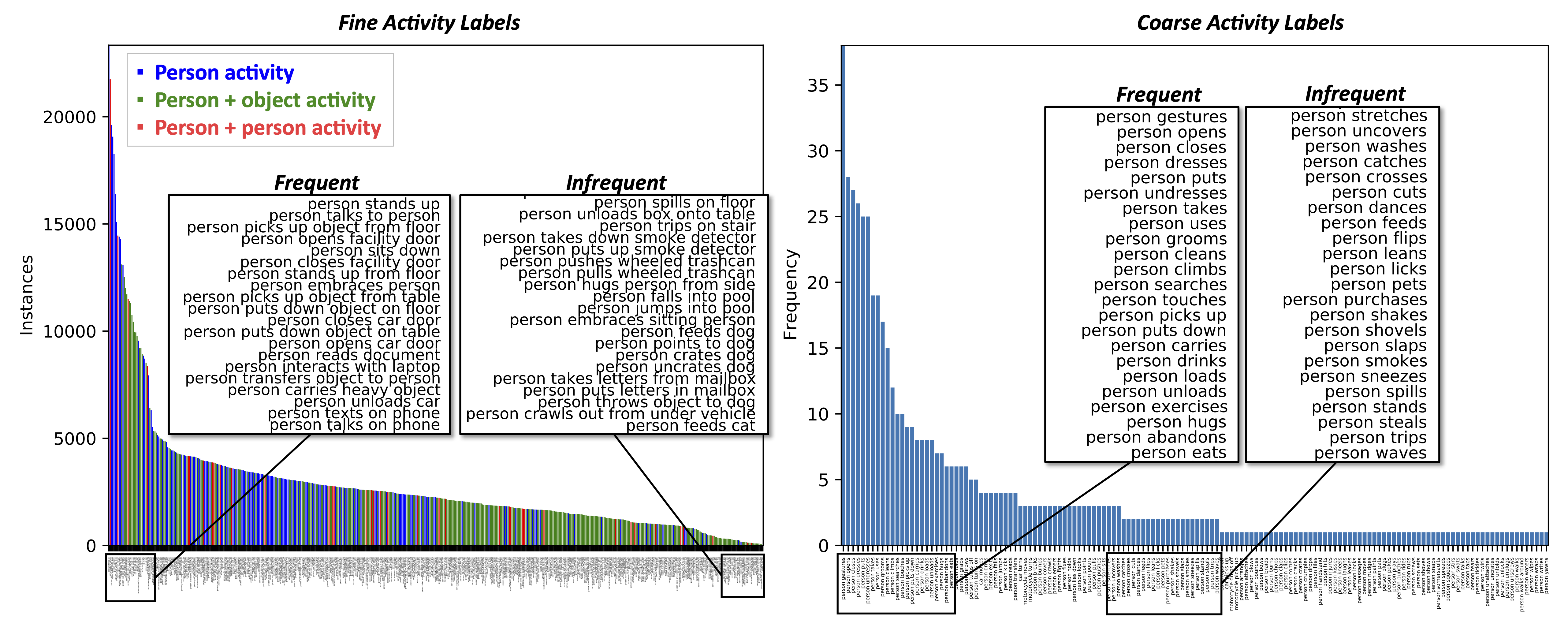} }
{\includegraphics[width=\figwidth]{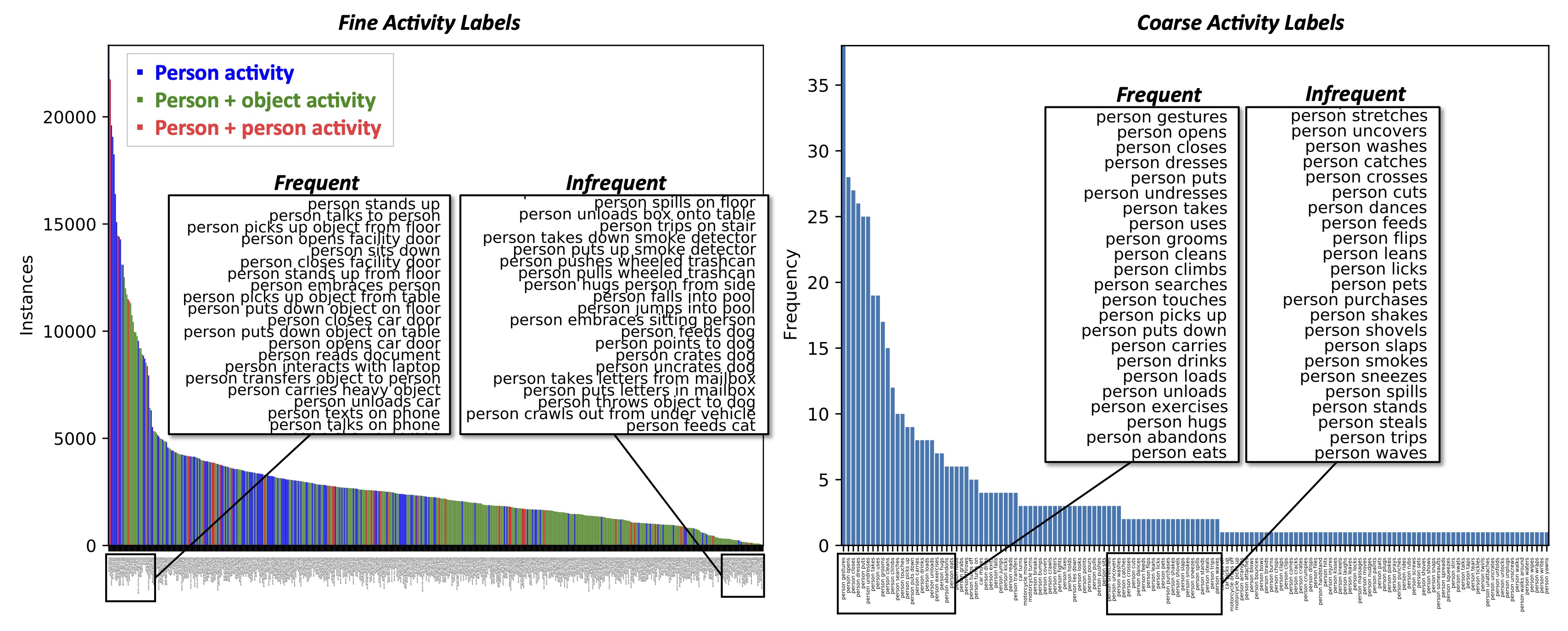}}
\caption{CAP label distribution.  (left) Instance histogram for fine-grained categories, colored by person-only, person-person or person-object interactions showing the most and least common labels by frequency, (right) Fine-grained histogram for each coarse-grained category to show the number of fine-grained categories in each hierarchical grouping.  Figure \ref{f:cap_labels} shows the full hierarchical label set.}
\label{f:cap_histogram}
\end{figure*}

\section{Consented Activities of People Dataset}
\label{s:cap}

The Consented Activities of People (CAP) dataset is a fine grained visual dataset of the activities of daily life, curated using the Collector platform.  Humans perform a wide variety of closely related activities every day that are subtle, localized and socially informative.  The CAP dataset was designed to explore the problem of representation of simple, fine-grained activities and provide a benchmark to characterize performance for classification and detection of these closely related activities. 

How do we define the set of labels in a dataset of fine-grained activities?  What exactly is a fine-grained activity?  The discussion of this question in appendix \ref{a:cap_design_challenges} suggests that a fine-grained activity is defined relative to other activities and should specify the following:

\smallskip
\begin{squishlist}
\item {\bf Who?} Fine-grained activity labels should be performed by the same noun (e.g. Person).
\item {\bf What?} Fine-grained activity labels should include simple verbs that can be performed in a few seconds along with other ``closely related'' verbs.
\item {\bf With?}  Fine-grained activity labels should include object interactions that induce a visually distinct motion.
\item {\bf How?}  Fine-grained activities should include visually grounded styles as within class variation.
\end{squishlist}

\medskip \noindent {\bf Label expansion}.  In order to downselect labels that satisfy these criteria, we perform a new strategy called {\em label expansion}.  Label expansion starts from the source labels in AVA \cite{gu2018ava}, Charades \cite{Sigurdsson2016}, Moments in Time \cite{Monfort2020}\cite{9609554}, Kinetics-700 \cite{kay2017kinetics}, Something-Something \cite{Goyal2017} and MEVA \cite{Corona2021WACV}.  We augment this set with the Activities of Daily Living \cite{adl}\cite{iadl}.  Next, we remove activities that are complex, non-visually grounded, non-person centered, not commonly performed around the house, or require skilled execution.  The remaining verbs are label expansion candidates.

We perform label expansion by selecting one or more closely related verbs and nouns for each label candidate that satisfy the CAP design goals.  We are all experts when it comes to understanding the subtle discrimination between gestures, social interactions or simple activities that we perform every day.  Therefore, the collector team leveraged their social expertise to manually perform label expansion for each candidate label.  For example, closely related verbs 
{\em person puts on socks} to {\em person puts on shoes} or closely related object interactions with different appearances {\em person puts on shoes} to {\em person puts on hat}.  

The result of the label expansion is is shown in figures \ref{f:cap_histogram} and appendix \ref{f:cap_labels}.   Appendix figure \ref{f:cap_labels} shows a circular tree plot of the hierarchical organization of the fine-grained labels grouped by ``Noun Verb'' structure, such as {\em person dresses} or {\em person gestures} into a two level, tree structured hierarchy.  

\medskip
\noindent {\bf Collection Campaign}.  The CAP campaign was set up to run on the Collector platform during the period of Apr 2020 to Dec 2021.  The CAP dataset was collected in two stages, Apr 2020 - Mar 2021 which focused on collection of MEVA activity classes \cite{Corona2021WACV} and July - Dec 2021 which focused on the remaining CAP activity classes.    The campaign specification includes 842 unique collection types, each specifies one of 512 activity labels and 157 object types.  In total, 288/842 collection types were specified so that the subject is facing away from the camera to increase diversity, 87/842 collections were specified to be collected to support temporal activity detection and 38/842 collection types were physically stabilized.
The overall collection statistics are shown in figure \ref{f:cap_statistics}, such that 905,369 clips are for activity classification (AC)  train/val, 132,271 clips for AC sequestered test and 416,900 clips for activity detection (AD).  Figure \ref{f:cap_histogram} shows the overall label frequency.  Note that this histogram is unbalanced due to frequent  organic activities, such as {\em person sits down} which often precedes object interactions.

The appendix discusses the key challenges (\S\ref{a:cap_design_challenges}), dataset design goals (\S\ref{a:design_objectives}), collection methodology (\S\ref{s:cap_collection}),  distribution format (\S\ref{a:cap_format}) and visualizations (Figure \ref{f:cap_explorer_large}, \ref{f:cap_ad_montage}) for curating a large scale dataset of daily activities.

\begin{figure*}[t!]
\centering
\includegraphics[width=\figwidth]{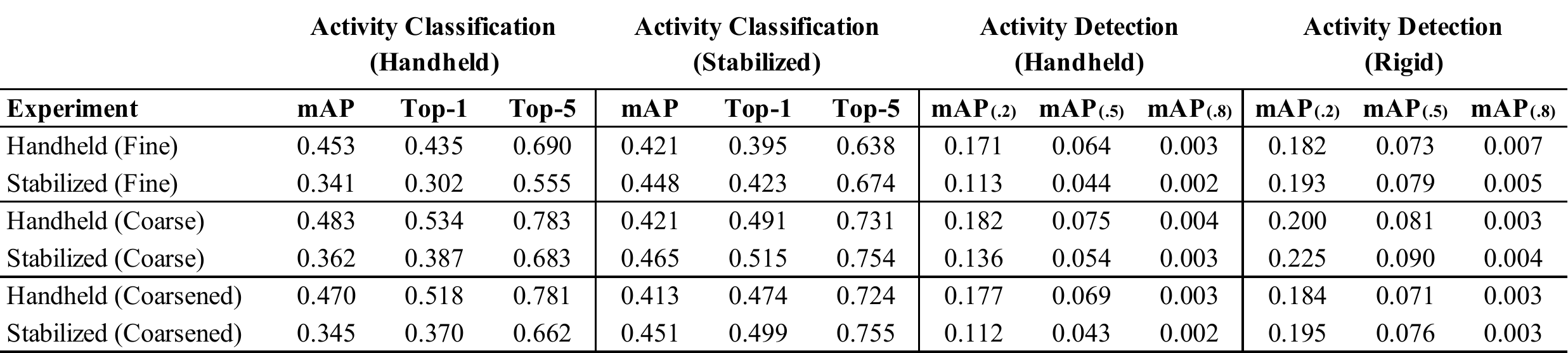}
\caption{CAP Benchmark Evaluation.  This result shows the performance of six experimental systems  (rows) on four evaluation tasks (columns).  The experimental systems differ in the training set, such that Handheld$\vert$Stabilized refers to the handheld or background stabilized video data and Fine$\vert$Coarse$\vert$Coarsened refers the training set labels (e.g. Fine labels, Coarse labels, or Coarsened labels trained on fine labels, then transformed to coarse labels at test time).  The evaluation tasks are Activity Classification$\vert$Activity Detection (\S\ref{s:tasks}) evaluated on Handheld$\vert$Stabilized$\vert$Rigid video subsets (e.g. handheld, software background stabilized or rigidly mounted video).}
\label{f:cap_results}
\end{figure*}

\section{Benchmark Suite}
\label{s:benchmark}

Performance benchmarking is the specification of an evaluation methodology, task, dataset and a baseline system design to evaluate system performance.  
Typical benchmarking considers test data that is in-domain, meaning it is collected and annotated exactly as it will be used in practice.  
However, consider the challenge of benchmarking fine-grained activity recognition in third-person security video.  We may collect many hours of video from many security cameras, without ever collecting an organic instance of a fine-grained target label like {\em person puts down backpack}.  If our goal is to benchmark performance for rarely occurring activities, then how do we benchmark in practice when the labels to evaluate almost never occur?  

We address this key challenge by introducing {\em domain adjacent benchmarking}.  In this strategy, we collect test sets that are from the required viewpoint, but with actors performing the test activities in short bursts.  This provides performance evaluation of a target domain (e.g. third person, long duration videos, organic activities) in a closely related adjacent domain (e.g. third person, short duration videos, actors).  The test data in the adjacent domain can be collected and distributed ethically, and performance evaluation on the domain adjacent data is used as a surrogate for the target domain.
Further discussion of the implicit biases in this strategy is provided in appendix \ref{a:benchmark}.

\subsection{Evaluation Tasks}
\label{s:tasks}

\noindent {\bf Activity Classification (AC)}.  The Activity Classification (AC) task is to assign one or more activity class labels and confidence scores to each video clip from a set of predefined classes.  The metric for AC performance is Mean Average Precision (mAP), top-1 and top-5 classification performance averaged over all classes.   

The AC task is separated into two domains, AC (Handheld) and AC (Stabilized).  AC (Handheld) is constructed using videos collected from handheld cameras, and AC (Stabilized) is constructed by performing software background stabilization on AC (Handheld) videos.  Appendix \ref{a:cap_format} discusses this background stabilization algorithm with examples shown in figure \ref{f:collector_post_processing_stabilization}.  The stabilization is used as a post-processing step to evaluate the domain mismatch of stabilized videos to rigidly mounted cameras.  

Figures \ref{f:collector_post_processing_refinement} and \ref{f:collector_post_processing_stabilization} show examples from the training set for the activity classification task.  The videos show untrimmed clips which include repetitions of an activity performed multiple times in a row by a subject.  The objective of the activity classification task is to specify a label for a three second trimmed clip containing one activity.

\medskip
\noindent {\bf Temporal Activity Detection (AD)}.  
The Temporal Activity Detection (AD) task is to detect and temporally localize all activity instances in untrimmed video. 
The metric for AD performance is Mean Average Precision (mAP) at a fixed temporal intersection over union (IoU) of 0.2, 0.5 and 0.8.

The AD Task is separated into two collection domains, AD (Handheld) and AD (Rigid).  AD (Handheld) is constructed from handheld cameras, and AD (Rigid) is constructed from rigidly mounted, unmoving cameras.  This separation is designed to evaluate a system trained with software stabilization, and tested on rigid cameras.  

Appendix figure \ref{f:cap_ad_montage} shows eight sample videos in the activity detection task.  This visualization shows seven frames extracted from a video on each row.  Each video is from a specific collection scenario, as described in section \ref{s:cap_collection}.  Each scenario has a subject performing between 7 and 11 activities in a sequence that is chosen by the subject.    

\begin{figure}[t!]
\centering
\includegraphics[width=\fighalfwidth]{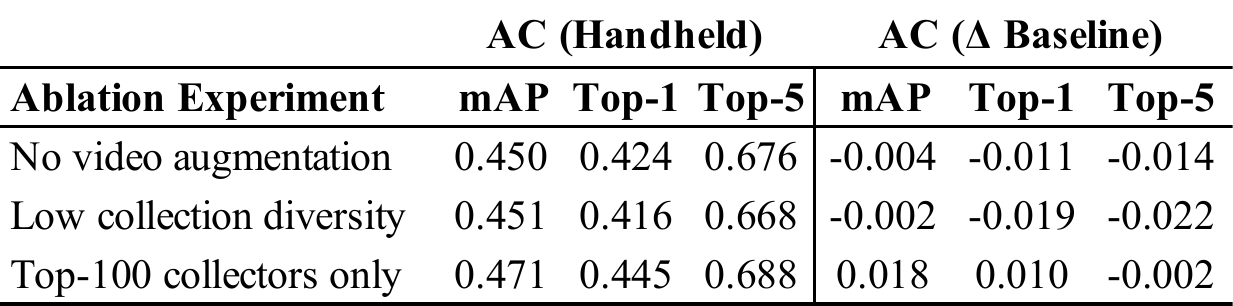}
\caption{CAP Ablation Study.  We retrained the baseline system removing video augmentation, removing the ``from behind'' collection diversity or removing all but the top-100 collectors, then compared the relative performance to the baseline.}
\label{f:cap_results_2}
\end{figure}

\subsection{Baseline system}

The baseline system for activity detection is based on activity classification of tracked cuboids \cite{Kalogeiton2017TubeletDF}.  The system operates by performing SORT tracking \cite{Bewley2016_sort} of people and vehicles, using a framewise YOLO-v5 \cite{Redmon2016YouOL} object detector on 5Hz videos followed by spatiotemporal IoU track association.  For each track above a minimum length ($>1s$) and minimum confidence ($>0.2$), define an activity cuboid proposal as the spatiotemporal sequence of bounding boxes for the object instance.  The cuboid is split into three second proposals, with overlap (4 frames), with replicated boundary conditions for short tracks, dilated by a constant factor (1.2), cropped to maximum square shape preserving the centroid and resized to 16x4x224x224 (frames, channels,height, width).  The cuboid is converted into a RGBA representation, with an alpha channel (A) encoding a binary mask for the tracked bounding box within the cuboid. Finally, we classify each cuboid proposal using a 3D-Resnet-50 \cite{hara3dcnns} with softmax classification  followed by a non-maximum suppression at temporal IoU $\ge 0.5$.  

Baseline training is performed using uniform random weight initialization, cross-entropy focal loss \cite{Lin2017FocalLF}, on 8 GPUs with minibatch size 256, ADAM optimization \cite{Kingma2015AdamAM} and inverse class frequency instance weighting on CAP dataset until validation loss saturates.  Data augmentation includes spatial mirroring and random clips by shifting $\pm 3$ frames.   
The baseline system is GPU optimized, real-time, python only and available at
\iftoggle{doubleblind} {\href{https://}{{\bf REDACTED}}}{\href{https://github.com/visym/heyvi}{github.com/visym/heyvi}}.

\begin{figure*}[t!]
\centering
\includegraphics[width=3.94in]{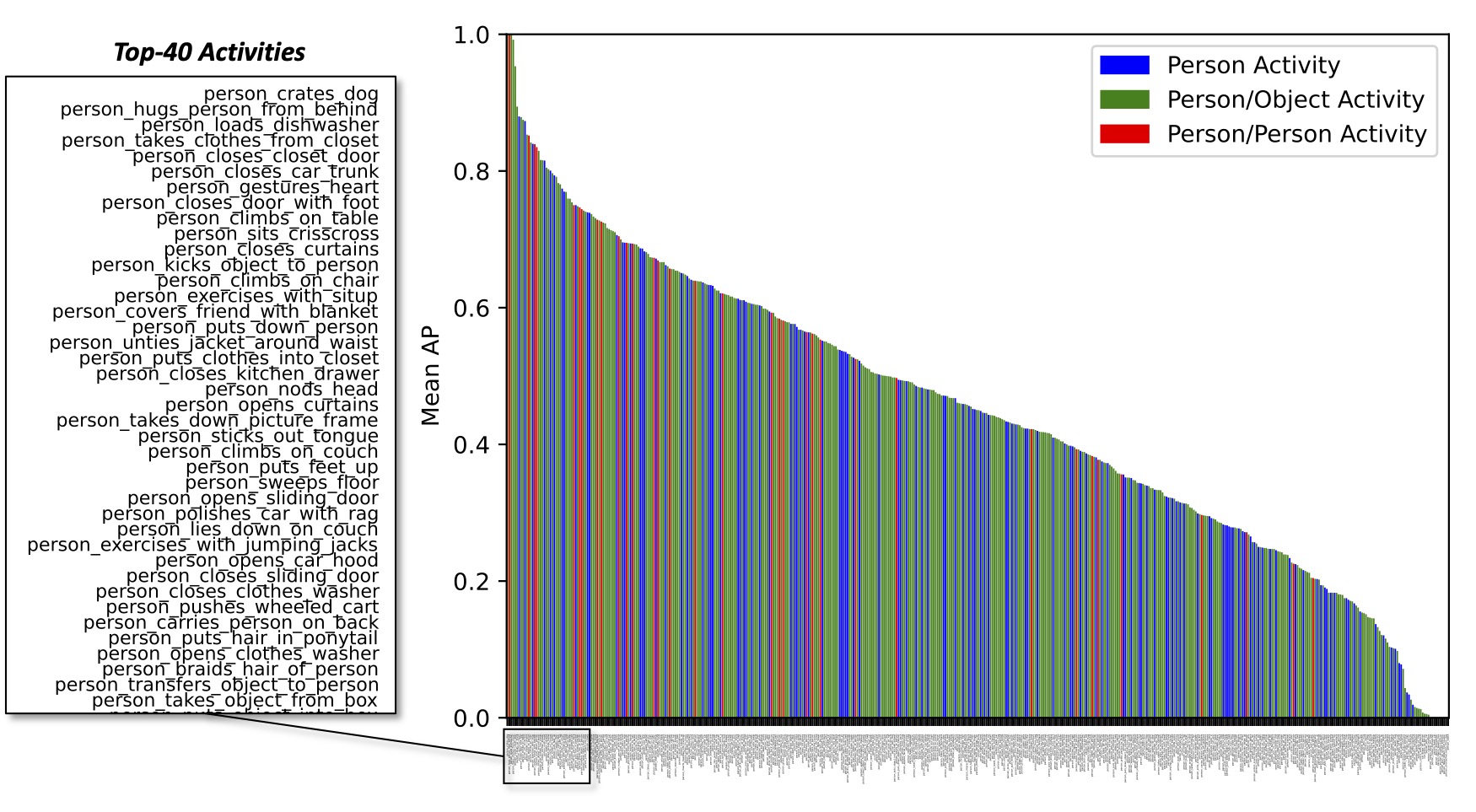}
\includegraphics[width=2.9in]{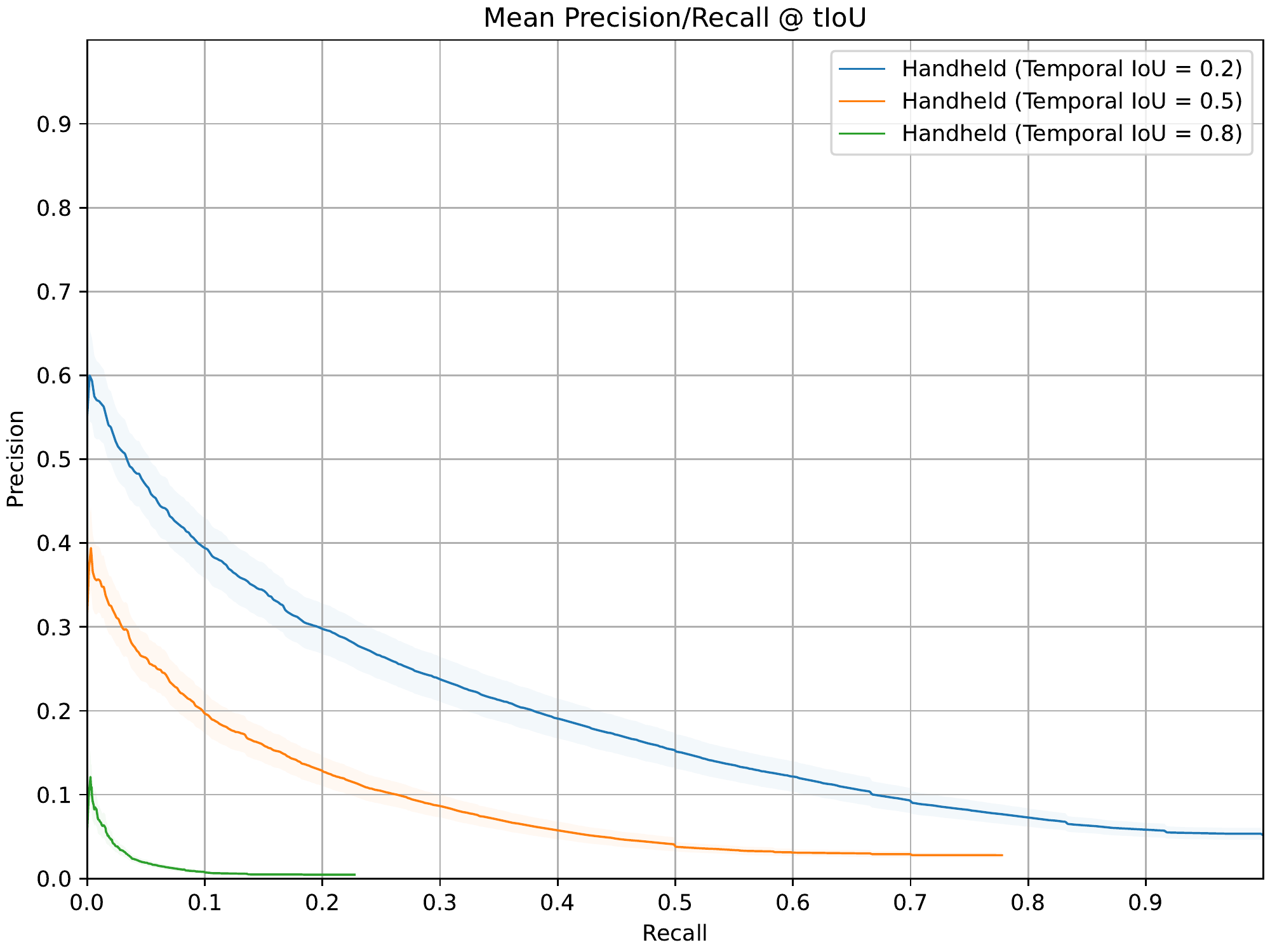}
\caption{CAP Benchmark Evaluation Plots. (left) Activity Classification (Stabilized) performance per class, sorted in decreasing order by mean AP, colored by person, person/object or person/person interactions, showing top-40 classes (zoom into PDF for rest), (right) Activity Detection (Handheld) performance showing the mean precision recall at ground truth assignment IoU=0.2, 0.5 or 0.8, with $1\sigma$ error bars.}
\label{f:cap_results_3}
\end{figure*}

\begin{figure}[t!]
\centering
\includegraphics[width=\fighalfwidth]{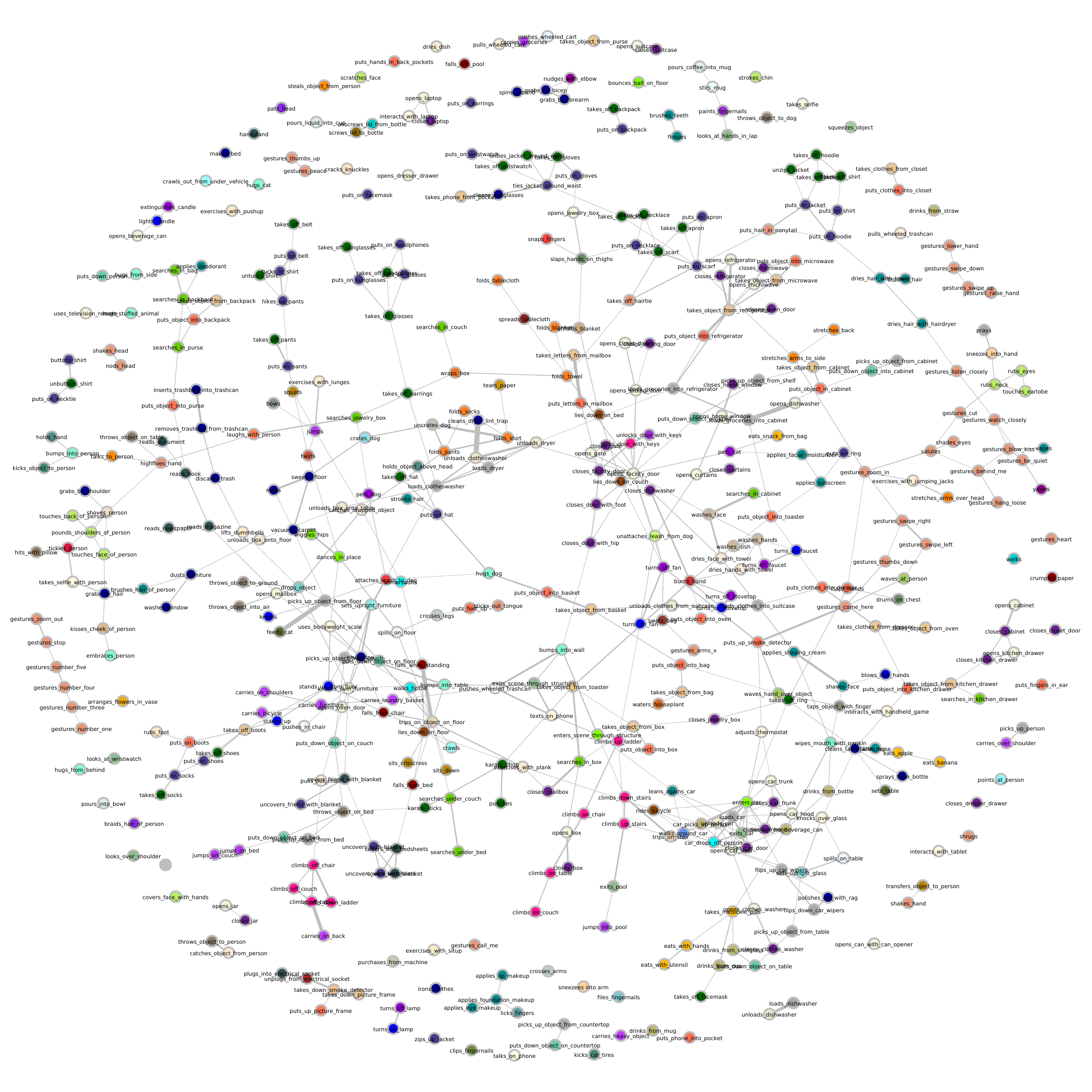}
\caption{Confusion graph for activity classification showing edges connecting commonly confused fine-grained activity labels.}  
\label{f:cap_benchmark_confusion_graph}
\end{figure}

\section{Performance Evaluation}
\label{s:results}

In this section, we describe the benchmark results on the CAP dataset, and results on the Activities in Extended Video (ActEV) Sequestered Data Leaderboard (SDL) \cite{Corona2021WACV}.

\subsection{Benchmark Results}
\label{s:cap_benchmark}

The experimental system runs the baseline with the following six combinations:  Handheld, Stabilized or Rigid videos with Fine, Coarse or Coarsened label sets.  {\em Handheld} refers to videos recorded directly from handheld mobile devices, {\em Stabilized} are the handheld videos with software background stabilization, and {\em Rigid} are test subset collected using rigidly mounted cameras.  The baseline system is trained using either handheld or background stabilized videos, with Fine labels (e.g. figure \ref{f:cap_labels} outer), Coarse labels (e.g. figure \ref{f:cap_labels} inner), or {\em Coarsened} labels which is trained using fine labels, then mapped via lookup to the coarse label at test time.  The benchmark datasets are the AC or AD sequestered test sets, subdivided into AC (Handheld), AC (Stabilized) and AD (Rigid) video subsets.

Figure \ref{f:cap_results} provides a benchmark evaluation on the activity classification and activity detection tasks.  
Results in figure \ref{f:cap_results} show: (i) background stabilization training helps for AD (Rigid), (ii) stabilized vs. handheld training exhibits a domain bias, (iii) AC (Handheld) performance is slightly better than AC (Stabilized) due to minor stabilization artifacts, (iv) AD is significantly more difficult than AC and (v) coarse labels are better than coarsened.

Figure \ref{f:cap_results_3} (left) shows activities on the AC (Stabilized) task ranked by mean AP per class and colored by object interaction type.  This provides a deeper insight into the classes that are the highest and lowest performing.  Results show that the best performing classes still leverage scene context (e.g. {\em crates dog, loads dishwasher}) and worst performing classes (mAP=0) and are poorly represented using the baseline system (e.g. {\em puts up smoke detector}).  Figure \ref{f:cap_results_3} (right) shows an aggregate result on the AD (Handheld) task, which demonstrates that fine-grained activity detection at precise temporal localization (IoU=0.8) is challenging.

\subsection{Benchmark Analysis}

Figure \ref{f:cap_results_2} shows the results of an ablation study to understand the effect of three training set configurations on baseline performance.  In all experiments, we renormalized the inverse class frequency weighting for the revised trainset, retrained the baseline system, then used the revised valset for model selection. First, we removed only the video augmentation (e.g. collectors performing activities multiple times), preserving all other data augmentation.  Results show that relative baseline performance is lower, which demonstrates that video augmentation helps.  Next, we removed only the ``from behind'' collections introduced for diversity.  Relative performance for this trainset is lower, which shows that collection diversity helps.  Finally, we kept only the videos from the top-100 collectors, comprising 65\% of training set.  Relative performance for top-100 trainset is higher, which suggests that for fine-grained activities (and our baseline system), it is better to have 
each collector perform many fine-grained activities.

Figure \ref{f:cap_benchmark_confusion_graph} shows a confusion graph of the AC task.  This visualization shows a 2-d graph embedding constructed by transforming a confusion matrix to a graph adjacency matrix such that nodes are fine grained activity labels, node colors are coarse grained labels, and edge thickness corresponds to commonly confused fine-grained activities.  A larger version is shown in appendix figure \ref{f:cap_benchmark_confusion_graph_large}.       

Analysis of the confusion graph provides four insights.  First, casual pairs (e.g. open and close) are commonly confused, since causal pairs often co-occur in a short temporal sequence.  
Second, we observe approximately one fourth of labels are not significantly confused, as shown by disconnected nodes.   
Third, there are small connected components with long range connections for common activities performed in sequence, such as interacting with drawers and cabinets in a kitchen.  
Finally, the three nodes that are most confused are {\em person trips on object on floor}, {\em person enters car} and {\em person opens facility door}, which suggests that improvement on these high degree labels should be prioritized.

\subsection{ActEV SDL}
\label{s:actev}

The ActEV SDL is a sequestered data leaderboard for activity detection in long duration security videos.  The ActEV SDL is labeled using the MEVA label set \cite{Corona2021WACV}, which include 37 simple activities in security video. The MEVA labels are a subset of the CAP labels with five additional labels for vehicles turning, stopping and starting.  We split the CAP dataset into a CAP-MEVA subset containing only the MEVA labels, which contains 405,781 background stabilized clips, split into 370K/35K train/val set.  CAP-MEVA was used to re-train the baseline system, and compared results to training using MEVA only, which contained 35,022 training clips, as of when this analysis was performed.

\begin{figure}[t!]
\centering
\includegraphics[width=\fighalfwidth]{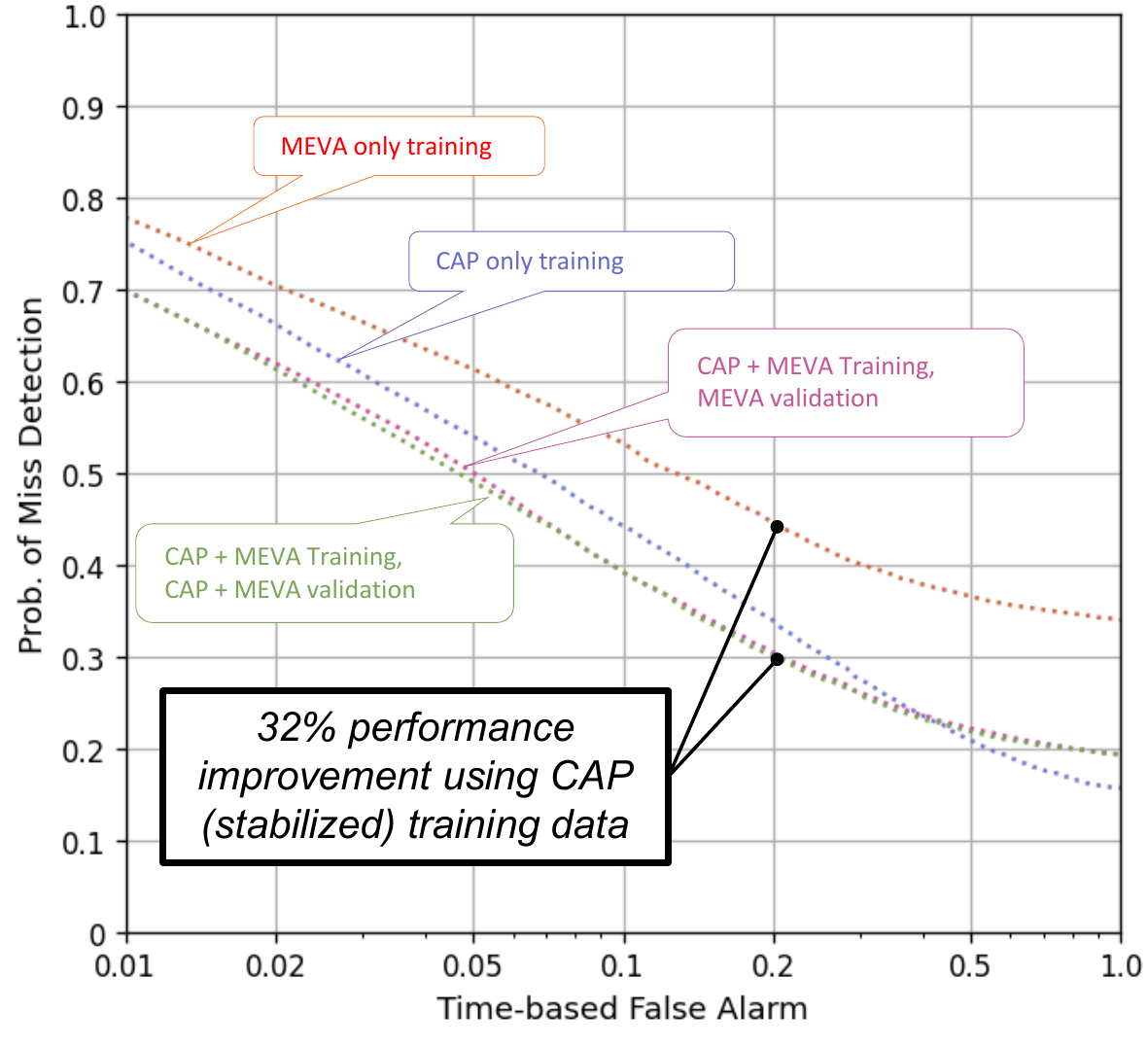}
\caption{ActEV SDL Evaluation.  (red) Trained with MEVA data only, (purple) trained with  the union of MEVA and a CAP  subset containing MEVA labels. Comparing the red/purple curves shows a 32\% improvement using CAP data for identical systems.}
\label{f:pip_370k_evaluation}
\end{figure}

Figure \ref{f:pip_370k_evaluation} shows an evaluation result on this dataset.  The performance metric is mean probability of missed detection over activity classes vs. time based false alarm rate (TFA).  We trained the baseline system using the MEVA dataset only or the union of CAP-MEVA and MEVA.    We made four submissions to the ActEV SDL that differed only by the training set and validation set assumptions.
Results show a 32\% improvement at a fixed TFA=0.2 due only to training with the CAP-MEVA data, when controlling for the training hyperparameters and system configuration.  Both green (MEVA + CAP-MEVA training) and red (MEVA only training) were trained from scratch rather than fine-tuned starting from a pretrained model. All training data is background stabilized.  This result shows that when controlling all other hyperparameters, the CAP dataset improves  sequestered temporal AD performance in long duration video.  This provides an independent validation of the CAP data for activity detection on static, long duration security video. 

\section{Conclusions}

In this paper, we introduced the Consented Activities of People dataset, the largest fine grained activity dataset of people ever collected.  Our benchmark provides a standardized evaluation of this new problem, with analysis to highlight the unique challenges of representing fine-grained activities.  
Finally, we believe that the Collector platform may be useful for the research community to address the never-ending demand for more ethical visual data.
 
\medskip
\noindent {\bf Acknowledgement.}  
\iftoggle{doubleblind} {{\bf REDACTED}.}{
Supported by the Intelligence Advanced Research Projects Activity (IARPA) via Department of Interior/ Interior Business Center (DOI/IBC) contract number D17PC00344. The U.S. Government is authorized to reproduce and distribute reprints for Governmental purposes notwithstanding any copyright annotation thereon. Disclaimer: The views and conclusions contained herein are those of the authors and should not be interpreted as necessarily representing the official policies or endorsements, either expressed or implied, of IARPA, DOI/IBC, or the U.S. Government.  Dataset distribution supported by the AWS Open Data Sponsorship Program.}

{\small
\bibliographystyle{./wacv23_template/ieee_fullname}
\bibliography{janus,jebyrne,gcastanon, datasets, finegrained}
}

\clearpage
\appendix

\iftoggle{doubleblind} {
\noindent {\bf ``Fine-grained Activities of People Worldwide''} 

\noindent Supplementary material for review

\noindent WACV 2023 submission \#\wacvPaperID
}{
\iftoggle{arXiv} {
{\noindent {\bf \LARGE Appendix}}}
{
\noindent {\bf ``Fine-grained Activities of People Worldwide''} 

\noindent Supplementary material, WACV 2023

\bigskip

{\noindent {\bf \LARGE Appendix}}}}

\renewcommand{\thesection}{\Alph{section}}
\renewcommand{\thefigure}{\Alph{section}.\arabic{figure}}

\section{Collector Platform}
\label{a:collector}

Collector is a global platform for collecting large scale consented video datasets of people for visual AI applications. Collector is able to record, annotate and verify custom video datasets of rarely occurring activities for training visual AI systems. 
The Collector platform provides:

\smallskip
\begin{squishlist}
\item On-demand collection of rarely occurring activities.
\item Simultaneous video recording, annotation and verification into a single unified platform.
\item Touchscreen UI for live annotation of bounding boxes, activity clips and object categories.
\item Specification of required collection attributes such as pose, illumination, location or object interactions. 
\item IRB approved informed consent for ethical dataset construction with in-app face anonymization.
\end{squishlist}
\smallskip

In this section, we will describe the motivation and design of the Collector platform.  This platform includes a {\em mobile app} for collecting labeled videos sourced from freelancers around the world, a {\em campaign dashboard} for setup, control and monitoring of a world-wide collection campaign by a dataset administrator and {\em quality control} for human review and distributed consensus for annotation quality.   This platform was used for collection of the dataset in section \ref{s:cap}, and we believe will be useful for the research community to support future dataset collection.  

\subsection{Mobile App for Recording and Annotation}
\label{a:mobile_app}

The front end of the Collector platform is a free mobile app designed to streamline consenting, recording, annotation and verification from collectors around the world.  

Figure \ref{f:collector_workflow_phase3} shows an overview of the collector workflow.  Collectors are invited onto the platform, and they download the collector mobile app to their device.  Collectors are presented \emph{collections}  which are video collection tasks grouped by required objects (e.g. a car, a motorcycle) or locations (e.g. parking lot, dining room). Each collection specifies the requirements of the submitted video, which include required activities, objects, location, illumination conditions, actor pose and camera viewpoint.  Once a collector chooses a collection to record, they get consent from their subject, including a video recording to ensure that the person consenting is the person being recorded.  Collector recruitment requires proficient readers of English in order to provide us informed consent in this step.  Next, the collector watches an example video which shows a gold standard exemplar of the collection.  We use visual exemplars to bypass language issues and communicate an idea of what the collection should look like.  Finally, the collector records and annotates the video live using touch gestures on their device, corrects errors using an in-app annotation editor and submits the annotated collection for review. Annotations include bounding boxes around objects, object labels and start and end times for each activity in the collection, all collected while the video is being recorded.  The best submissions from our worldwide collection team as adjudicated by the review team are used as new training examples for newer collectors.  In other words, collectors ``see one, do one, and teach one'' on our platform.

\begin{figure*}[t!]
\centering
\includegraphics[width=\figwidth]{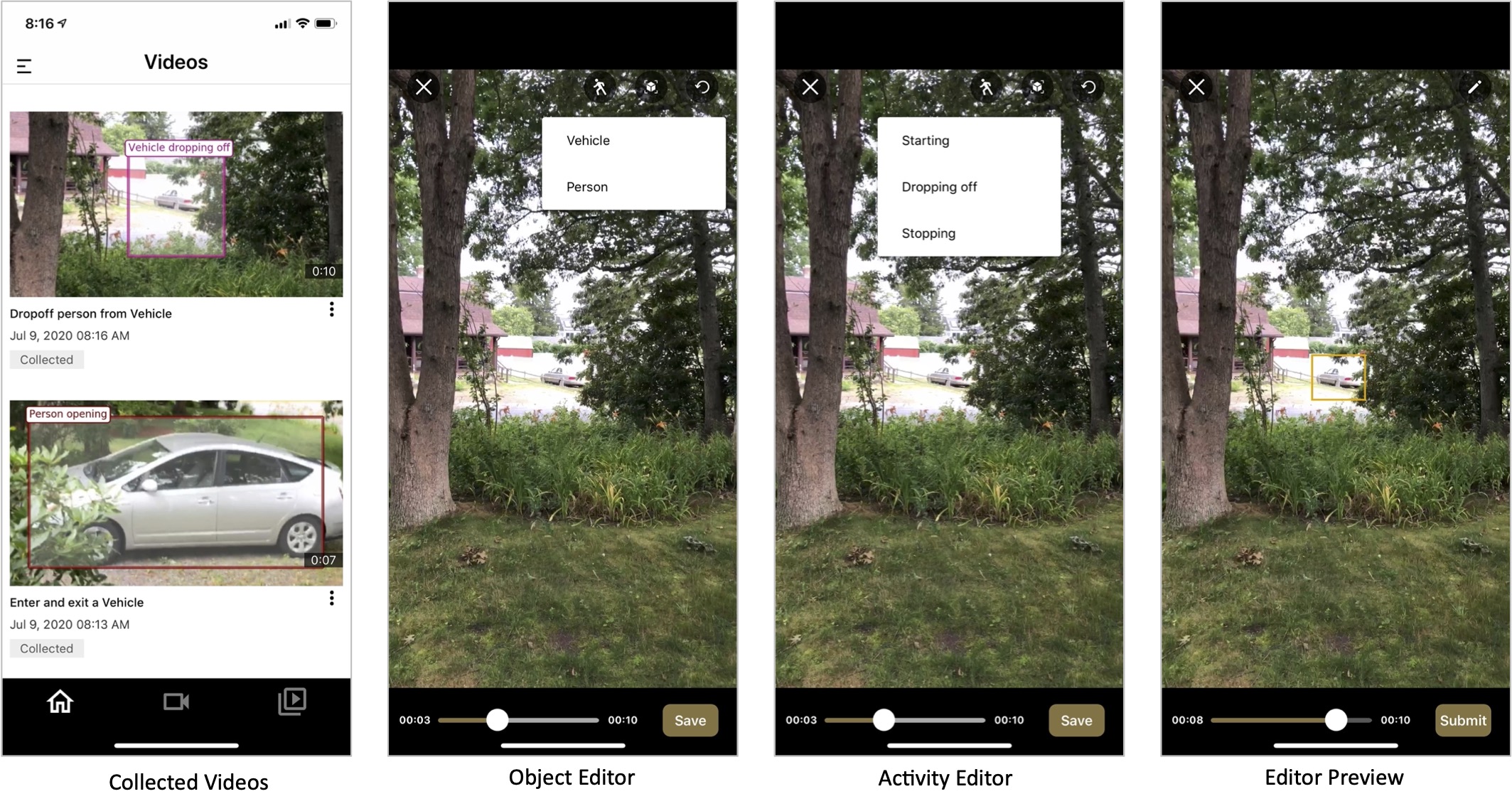}
\caption{Editing in the Collector mobile app.  The editor is an in-app tool that allows for modifying the bounding box and activity timing for objects in video.  This allows for correction of live annotations that were inaccurately collected at recording time.  Editing is performed by using multi-touch gestures to define bounding boxes that deform through time to track objects, and press gestures within boxes to define start and end times for activities.  A video tutorial for the editor is available at 
\iftoggle{doubleblind} {{\bf REDACTED}}{\href{https://visym.com/editor}{visym.com/editor}}.
}
\label{f:collector_editor}
\end{figure*}

Figure \ref{f:collector_editor} shows an example of the in-app annotation editor.  This editor is used to annotate videos in-app after they have been collected.  This is useful for collecting data to support the Activity Detection (Rigid) task, where the device must be stationary during recording, where collector cannot annotate in-app while they record.  Annotations include bounding boxes for objects and people, which is specified using multi-touch gestures for fast video annotation.  Further annotations include start and end times for activities, which are specified by press gestures in a bounding box for the start (press-down) and end (lift) when an activity occurs.  The saved edited video is uploaded to the Collector backend for further processing.

\begin{figure*}[t!]
\centering
\includegraphics[width=\figwidth]{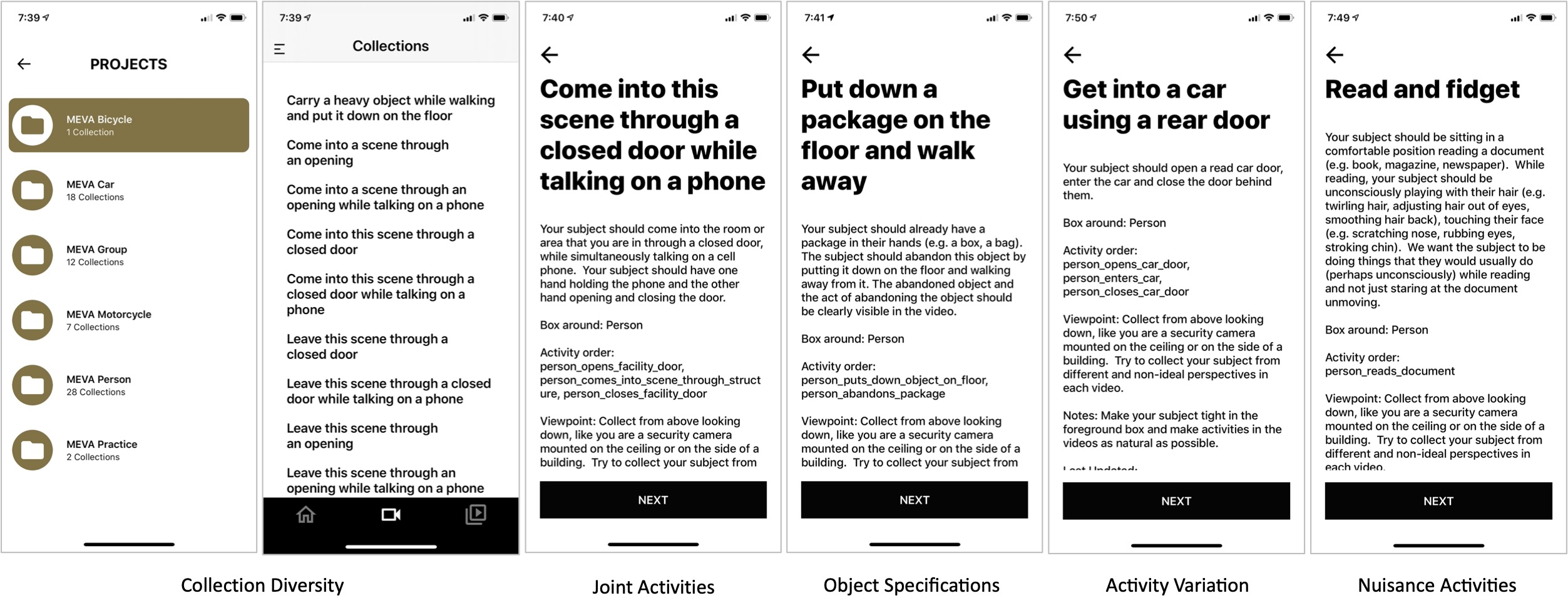}
\caption{Collection diversity on the Collector platform.  The platform allows for controlling the diversity of collections, by introducing collection variants with simultaneously occurring activities, new objects, within-class variation or rarely occurring activities.}
\label{f:collector_diversity}
\end{figure*}

Figure \ref{f:collector_diversity} shows an example of increasing the diversity of collections by controlling the collections available in the campaign.  These screenshots show what is presented to the collectors in-app when they are tasked with collecting diverse data.  The key component for this workflow is showing the collectors an example video for what should attempt to collect along with a written description.  This provides multiple resources to the collectors to aid them in collecting high quality data of the form needed by the campaign.

The Collector mobile app is freely available in the iOS and Android app stores.  This app has been downloaded and used by thousands of freelance collectors worldwide.  More information and a tutorial video for collector usage is available at
\iftoggle{doubleblind} {{\bf REDACTED}}{\href{https://visym.com/collector}{visym.com/collector}}.

\subsection{Campaign Dashboard for Global Coordination}
\label{a:campaign_dashboard}

Large scale dataset collection includes a large volume of videos to be collected, annotated and verified.  Each collection campaign may contain hundreds of different collection types, and each collection type may have tens of thousands of video submissions from all around the world.   Campaign management tools are critical to coordinate the workforce, monitor submissions for fraud or unclear instructions, and maintain high quality for this volume of data.

The collector platform coordinates a global workforce using a {\em campaign dashboard}.
The campaign dashboard provides a real-time interface for an administrator to set up, control and monitor a campaign.  The dashboard enables an administrator to:
\smallskip
\begin{squishlist}
    \item {\bf Plan}.  The admin defines the campaign by specifying each collection, along with the required objects, viewpoints, styles, illuminations, locations and IRB approved consent language. 
    \item {\bf Train}.  The admin team collects one or more reference videos for each collection.   Each collector is shown the description of a collection, and they view one or more reference videos for this collection.  This provides a ``see one, do one'' strategy for non-native English speakers.
    \item {\bf Deploy}.  The campaign is deployed based on a collection schedule, including a maximum number of collections allowed per collector and per campaign.  Admins can onboard and offboard collectors worldwide.
    \item {\bf Monitor}.  The dashboard shows a view live submission stream from collectors and reviewers worldwide. This allows the dataset admin to confirm that the collection is being submitted correctly and successfully.
    \item {\bf Refine}.  The dataset admin can refine the collection or training videos based on the performance of the the collector community.  For example, appendix figure \ref{f:collector_diversity} shows examples of refining a collection to increase diversity of objects and activities.
    \item {\bf Pay}.  Freelance collectors worldwide are paid per video that is submitted and accepted by the review team.  Payments are made on a bi-weekly basis.      
\end{squishlist}
%

\begin{figure*}[t!]
\centering
\includegraphics[width=\figwidth]{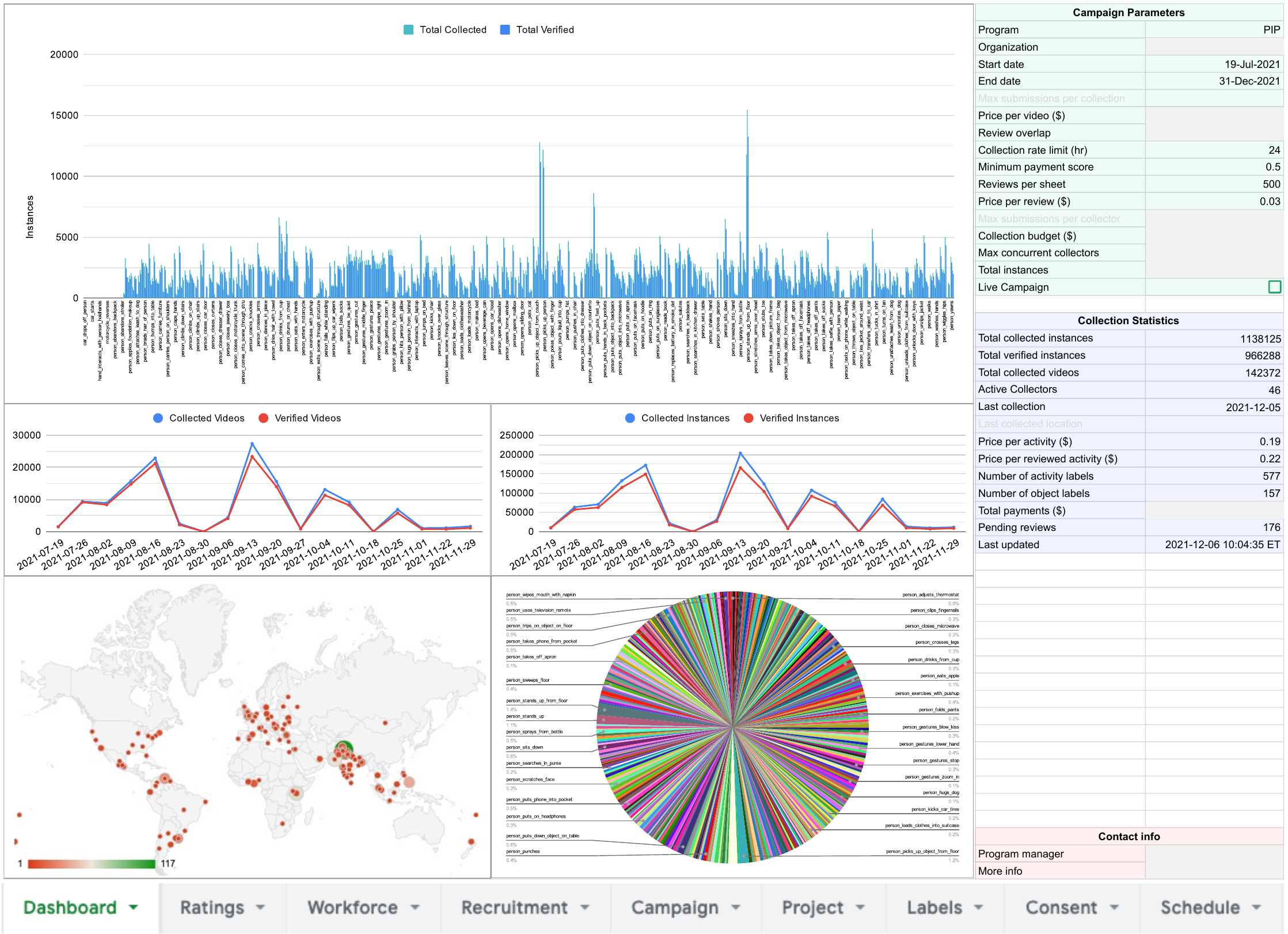}
\caption{Campaign Dashboard on the Collector platform.  The campaign dashboard is a live HTML interface used by a campaign administrator to monitor collection status, view live submission stream, list active collectors worldwide, modify collections in-app, define new labels and specify IRB consent language.  The bottom tabs expand into specific views of the campaign focused on Ratings, Workforce payments, Recruitment onboarding/offboarding and Campaign specification.  The dashboard shows the live histogram of total collected instances, collected and accepted videos by week, submissions by geographic distribution, and pie graph of labels submitted and total collection statistics.  Finally, the campaign parameters provide command and control of the collection campaign.
}
\label{f:collector_dashboard}
\end{figure*}

Figure \ref{f:collector_dashboard} shows an example of the live dashboarding providing the state of the campaign.  This dashboard is the primary visualization of the state of the collection campaign that is used by an administrator for monitoring, command and control. This captures a global near real-time view of the worldwide submissions along with the ability to visualize submissions from any collector or reviewer.

\subsection{Human Review for Annotation Quality}
\label{a:human_review}

Large scale dataset collection requires careful review of videos to maintain dataset quality.  In the Collector platform, the review team is tasked with daily verification of submissions to check that they satisfy the collection requirements.  Reviewers are promoted from within the worldwide collector pool, and are selected based on their historical submission quality, and their interest in reviewing the work of fellow collectors in exchange for a fixed price per review.

\begin{figure*}[t!]
\centering
\includegraphics[width=\figwidth]{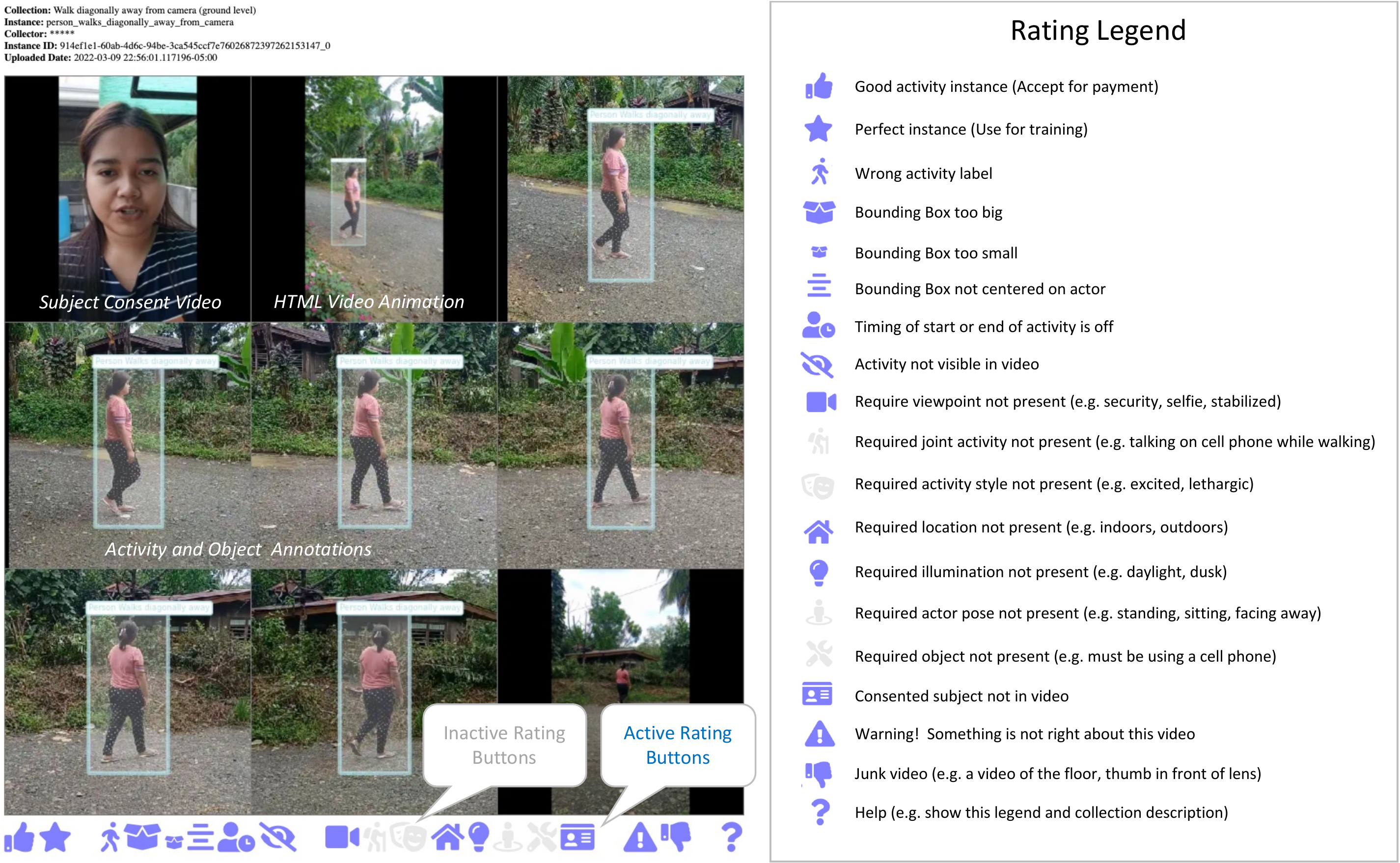}
\caption{Reviewing interface on the Collector platform.  Reviewers are provided an HTML interface for each video under review, that provides buttons for feedback.  The reviewer can click on the image to show a WEBP animation of annotated video, along with the consent video in the upper left to confirm that the recorded subject is the subject that provided consent.  Reviewers are tasked with selecting one of the review buttons at the bottom for this video, which are streamed to the Collector backend for aggregation.  The rating legend shows the description for each buttons, such that ``grey'' is disabled and is not required for this submission.}
\label{f:collector_review_interface}
\end{figure*}

Figure \ref{f:collector_review_interface} shows the reviewing interface.  These screen shots show what is presented with reviewers when they are tasked with maintaining quality control.  We have developed an HTML based interface for quickly reviewing the annotation quality of a video submitted by the collectors.   Reviewers are sent a daily email with an HTML review link containing their reviews for the day.  This review interface displays animated WEBP clips, in a montage format that allows for fast reviewing of a video at a glance.  Reviewers are tasked with (i) reading the collection description, (ii) watching the reference example video for this collection, (iii) watching the submitted video and (iv) selecting one or more reviews by pressing the appropriate HTML button.  

Each video is reviewed by a fixed number of reviewers (usually three), and a mean quality score is assigned to each video.  Reviews are streamed to the Collector backend for aggregation, and reviewers are paid a fixed fee for each review submitted.  Reviewers are audited by providing them reviews in their review stream with a known label (e.g. a synthetically corrupted video or a video reviewed by an administrator), to check their review quality and give them feedback.  Finally, if this video quality is above a campaign specified threshold, then it is authorized for payment.

\section{Consented Activities of People Dataset}
\label{a:cap}

In this section, we discuss the related work, key challenges, collection methodology and distribution format for curating a large scale dataset of activities of daily life. Section \ref{a:cap_design_challenges} specifies the design challenges of this dataset and motivates our design goals.  Section \ref{s:cap_collection} describes the collection parameters  using the Collector platform.  Finally, section \ref{a:cap_format} describes the format of the dataset and the evaluation tasks.

\subsection{Related Work}
\label{a:comparison}

Table \ref{t:comparison} shows a dataset comparison with the state of the art.  
For datasets with multiple evaluation tasks, we select the task and associated data most closely related to activity classification or activity detection.  For example, Ego4D has five benchmark tasks, and we compare with the subset of data labeled for the Moment Query (MQ) task.  For those datasets with label space organized as a multi-level hierarchy (e.g. ActivityNet) we select the lowest level as the number of fine classes and the immediate parents as the number of coarse classes.  If there is no hierarchical organization, we report the number of classes as coarse classes.  Clips reports the number of instances of each class (e.g. a trimmed clip containing an activity) available for testing, validation or training.  We show only those mean clips per class that are reported by the authors in the source publication.  The CAP dataset reports the mean clips per class across all classes, and mean clips per class considering only the top-250 fine-classes with the largest number of instances, as shown in figure \ref{f:cap_histogram} (left).  Finally, note that this analysis does not include specialized domains such as fine-grained activities in sports datasets  \cite{Deliege2020SoccerNetv2}\cite{Karpathy2014}\cite{FineAction2020}\cite{li2021multisports}\cite{1804.03247}\cite{6247801}\cite{2105.11107}\cite{2105.07404}.

This table shows that the proposed CAP dataset is the largest consented dataset of people as measured by mean clips per class for training.  

\subsection{Design Objectives}
\label{a:design_objectives}
The CAP dataset has the following design objectives:

\medskip
\begin{squishlist}
\item {\bf Atomic.} Activities should be short duration with length $\leq$ 3 seconds and visually grounded (e.g. activities should be discriminative from the pixels). 
\item {\bf Fine-grained.} Activities should be selected where motion is critical for discrimination, rather than the scene context or object appearance.  
\item {\bf Daily-life.} The collection should involve locations, activities and objects that people use or perform every day, without practice or expertise.
\item {\bf Non-overlapping.} All activities should be performed independently. (e.g. a subject will not simultaneously use a cell phone while taking off a hat).
\item {\bf Person-centered}.  All videos should include a primary consented person performing the activity. 
\item {\bf Third-person}.  All videos should be collected from a third-person viewpoint, looking down on the scene from above, consistent with a ceiling/wall mounted camera.  
\item {\bf Diverse}. Activities should be collected to encourage diversity in culture, geographic location, viewpoint, objects, pose and illumination.
\item {\bf Worldwide.} Videos should be collected from many countries around the world.
\item {\bf Ethical.} All videos must be collected with informed consent for how the videos will be shared and used. 
\item {\bf Balanced}. Activities should be collected so that the number of instances per class is approximately equal, including labels that are rare in natural video.  
\item {\bf Large-scale.} The dataset should include a liberal license with open distribution format and easily downloadable training and validation set.
\item {\bf Annotated}. Videos should be annotated with bounding box tracks around the primary actor along with temporal start/end frames for each activity instance. 
\end{squishlist}

\subsection{Design Challenges}
\label{a:cap_design_challenges}

These design objectives introduce a number of challenges and open questions.  What is a fine-grained activity?  How are the activity labels selected?  How can we collect balanced data of infrequent labels?  How do we control the diversity of the data collected?  How do we ensure a dataset is collected both globally and ethically?  

What are fine-grained activities?  Fine-grained visual categorization is an established task in the object recognition  literature \cite{Gao2016, Cui2017, Yandex2015, Rao2021, Moghimi2016, Yu2018, Lin2017, Kong2017, Chen2019}. 
The term ``fine-grained label'' or ``fine grained category'' was originally introduced in the context of image classification of subordinate object categories, such as bird species, plant species or product brands.  These are classes with subtle discriminative features, such as the color of a wingtip or shape of a leaf.  These annotations often require an expert to specify the class label, and the differences between classes are subtle, highly localized and require expert training.  However, the differences between classes are visually grounded in the pixels, if you know where to look.  

A fine-grained activity is not as straightforward to define.  An activity is typically defined by a verb being performed by a noun.  For example, the activity {\em person sits} has the noun ``person''  performing the verb ``sits''.  Is this a fine-grained category?  Compare the activity category of {\em person sits} to {\em person squats} vs. {\em person drinks}.  In the first case, there are subtle differences in how the lower body is moved in sitting by resting your body weight against a flat surface as compared to squatting down by bending your knees while resting your body weight on your feet.  In the second case, there are clear motion differences between drinking with your upper body as compared to sitting with your lower body.  This suggests that a fine-grained class requires subtle motion differences with other closely related verbs.   

Is a fine-grained activity defined by the noun performing it?   For example, compare the activity category of {\em dog sits} vs. {\em person sits}.  The kinematics of a four legged animal sitting on hind legs exhibits a different motion than a two legged person sitting in a chair.  Furthermore, the object appearance of the noun ``dog'' can provide context to aid in the recognition of {\em dog sits} as compared to {\em person sits}.  The discrimination of a fine-grained activity class should primarily require representation of the motion being performed and not exploiting object appearance or scene context cues.  This suggests that the set of fine grained activities should be performed by the same noun in order to remove the confounding effects of object or scene context.  This does introduce a challenge of combinatorial scale when composing noun/verb pairs, however a dataset focused on people only will avoid this combinatorial explosion. 

Is a fine-grained activity defined by an object being interacted with?  There exist scenarios with an actor interacting with visually distinct objects that exhibit the same or different motion pattern when performing the activity.  For example, consider the activities {\em person throws baseball} vs. {\em person throws rock}.  These activities exhibit largely the same throwing motion of either a rock or a baseball and the use of the object does not change the motion of the activity.  These motions are not visually distinct, and are only distinguishable through identification of the object category ``rock'' or ``baseball''.  However, consider {\em person carries bicycle} vs. {\em person carries groceries}.  Carrying a heavy bicycle is awkward and requires a different strategy of carrying over your shoulder or pulling the object towards your chest, as compared to lifting grocery bags with handles in either hand.  These are both examples of carrying a heavy object lifting using your upper body then walking, but the motion induced by the object when performing the activity is different.  This suggests that a set of fine-grained activities should include object interactions that induce visually distinct motions.  

Is a fine-grained activity defined by the style in which it is performed?  An activity style can be described in terms of an adverb that modifies the verb being performed, such as ``skillful'' or ``clumsy''.  We humans  are experts at subtle discrimination between gestures or social interactions and we are highly tuned to picking up on the body language in how an activity is performed.  However, the same visually grounded style can be performed for more than one activity, such as {\em person sits skillfully} as a dancer would or {\em person jumps skillfully} using an efficient economy of motion.  This suggests that style is an important attribute for within-class variation, but should not be considered as a separate class.

Finally, why do we need atomic fine-grained activities?  Our core hypothesis is that training a visual AI system that represents the diversity of human activities will result in a representation of subtle motions with improved generalization performance for real scenes.  Large-scale pretraining requires significant overlap between label set and target domain \cite{Mahajan18}\cite{Chen2021DeepAO}.  Our collection goals are to collect the activities of daily life from third person viewpoints to provide a pretraining and fine-tuning dataset for activity detection of daily life.    
For example, to detect when a person is talking on a cell phone, we can include closely related activities in training such as scratching your head or putting on headphones.  This forces a more precise learned representation of talking on a cell phone that is not always predicted simply when you touch your ear.  Furthermore, a training dataset that includes simple, short and atomic activities can provide a foundation for study of longer complex or composite activities of daily life that require reasoning about intent or identity.  Our goal is to explore this representational ability for atomic activities, and provide a foundational open dataset to explore ethical human activity detection.

\medskip
\noindent{\bf Long-tailed classes}.
Human activities are diverse and long-tailed.  There exist activities that each of us perform many times per day, such as standing up or sitting down, opening or closing doors or getting dressed.  These common activities are diverse in that there are many ways that one can perform each activity which change the appearance, such as sitting criss-cross on the floor vs. sitting in a chair, opening a sliding glass door vs. opening a facility door or putting on a hat vs. putting on a jacket.   These within-class variations of activities are in addition to the more common variations due to camera pose, actor pose or illumination.  
These within-class variations of activities are {\em diverse} which capture the variability of naturally occurring human activities.

\begin{figure*}[t]
\centering
\includegraphics[width=\figthirdwidth]{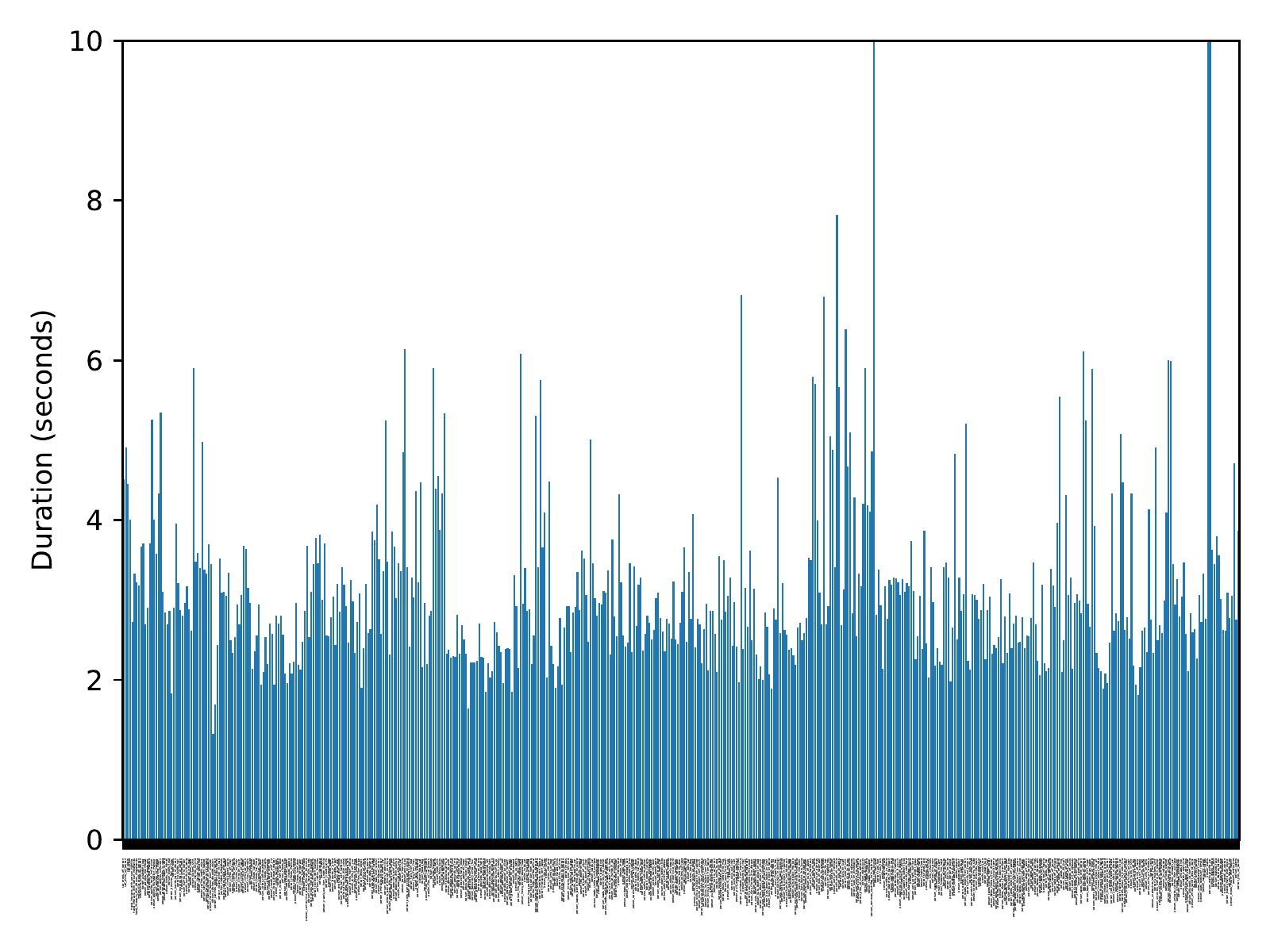}
\includegraphics[width=\figthirdwidth]{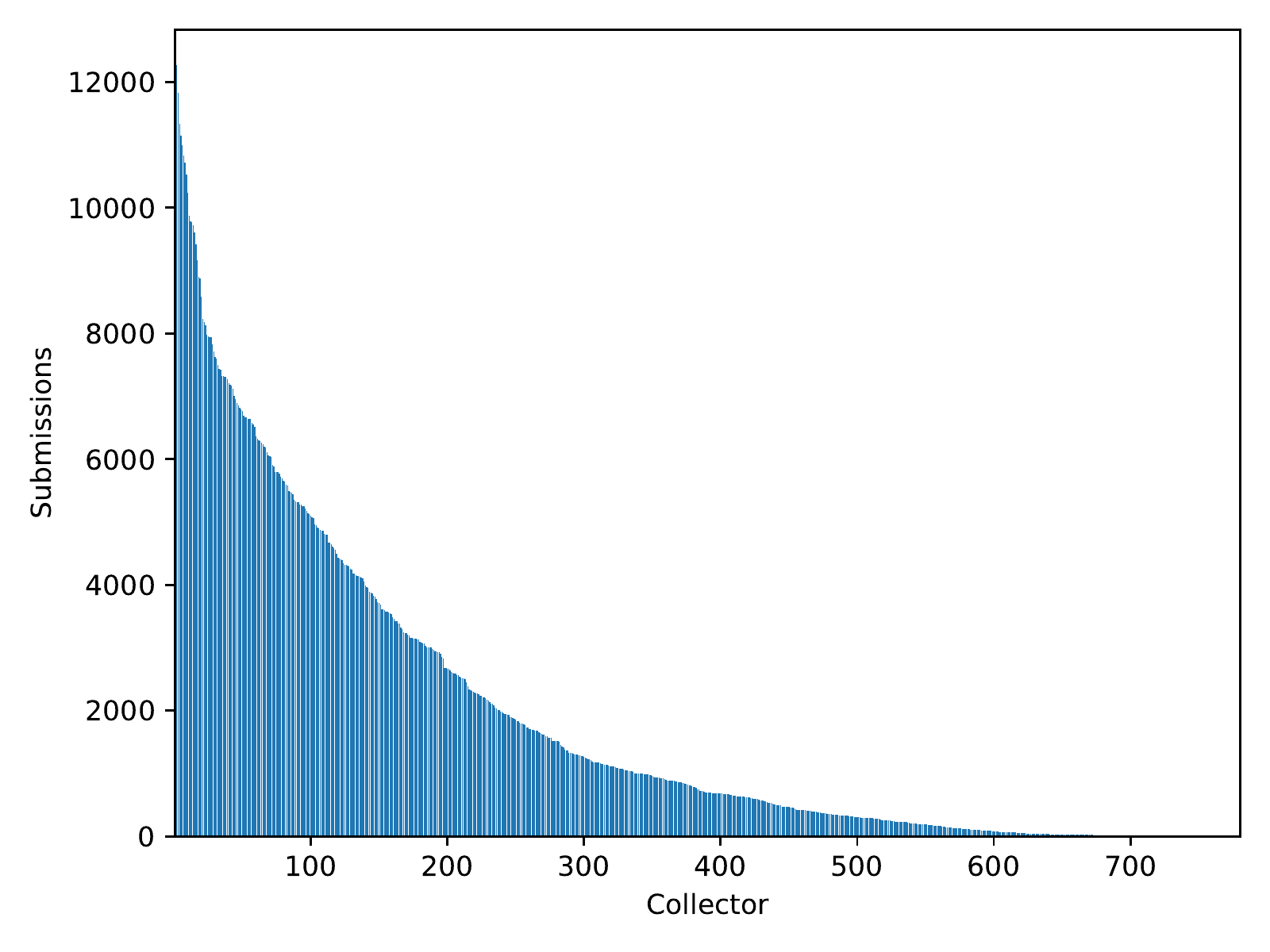}
\includegraphics[width=\figthirdwidth]{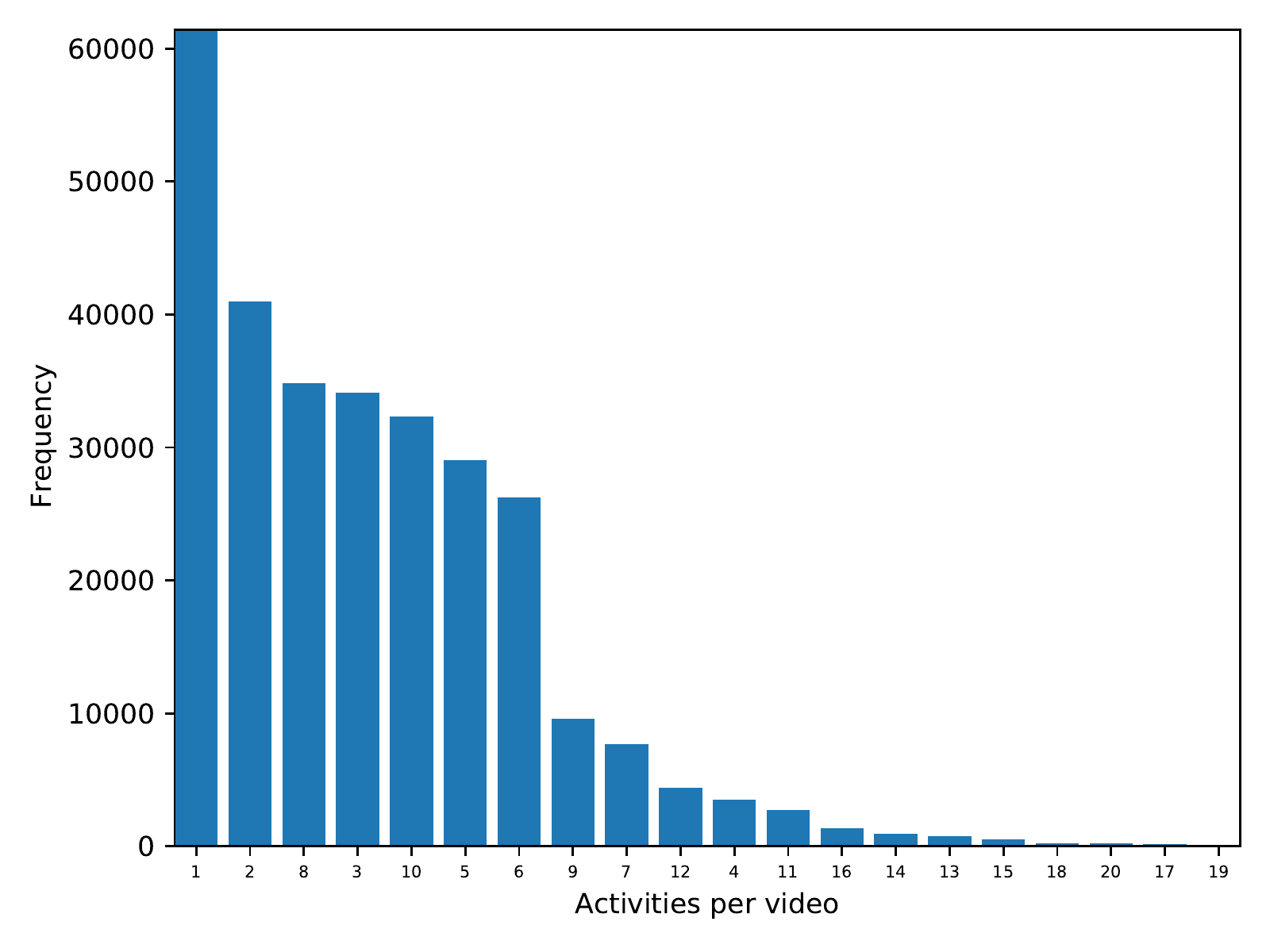}
\caption{CAP Dataset statistics.  (left) Mean duration in seconds of each activity label, (middle) The number of submissions for each collector sorted by maximum number of submissions which shows that some collectors are enthusiastic about making contributions, (right) The sorted density of the number of activity instances per video, which shows that the most common number of activity instances per video are 1, 2 or 8 instances.  We recommend zooming into the PDF to see the activity labels.}
\label{f:cap_details_1}
\end{figure*}

Human activities are also {\em long-tailed}.  Just as there are activities that each of us perform frequently, there are many more activities that we perform infrequently or possibly never.  For example, for some people this may be domestic activities such as cooking, cleaning or folding, for others it may be violent activities, such as fighting, others may be potentially harmful activities such as tripping or falling.  These are activities that may occur so rarely that in months of video (or scraping videos from social media) no examples are captured.  Furthermore, even if these activities occur, annotating them in long duration videos requires manual search through many hours of video to localize rare activity instances.  It becomes increasingly difficult for annotators to remember all activities that have been specified as the number of activity classes increases.  This imbalanced frequency distribution of the occurrence of human activities is long tailed in that there are fewer classes that are performed frequently, but many more classes in the tail of this distribution that are infrequent.  

\medskip
\noindent{\bf Importance sampling}.  How can we create a balanced dataset that includes both diverse and long-tailed human activities?  One strategy is to consider a video as a sample of the visual world, such that a video dataset contains a finite sample for a specific task.  Direct sampling (e.g. point a camera out the window) leads to a dataset that may be large scale, but imbalanced, containing frequently occurring labels (e.g.  people walking), but will under-represent rare and fine activities that may never occur here.  

Consider an alternative strategy of {\em importance sampling}, which samples videos given pre-selected labels.  This strategy generating samples on-demand for nearly any desired label.This introduces a tradeoff between direct sampling which curates videos that are naturally occurring and frequent (but imbalanced) vs. importance sampling that are engineered and balanced (but potentially biased).     

Section \ref{s:collector} introduced the Collector platform to address the issue of diverse and long-tailed classes by enabling {\em on-demand collection}.  The Collector mobile app enables activities to be requested, recorded, annotated and submitted on-demand from a global workforce of collectors.  These collection requests are controlled to balance out the frequency of long-tailed classes.  Furthermore, the collections break down activities into variations that capture the diversity of the activity class.  For example, to address the diversity of ``person dressing'' we can release a collection request for different variations of dressing: person puts on hat, person puts on shirt, person puts on belt, person puts on pants, person puts on socks, person puts on shoes, etc.  To address the diversity of illuminations, we can request a video in indoor illumination conditions or in darkened conditions.  This platform operationalizes the dataset collection strategy of importance sampling.  Section \ref{s:cap_collection} discusses this further.

\medskip
\noindent {\bf Domain Adjacency}.
Domain adjacency is the problem of collecting training and testing data in a given source domain, which will be deployed to a closely related (but not identical) target domain.  An ideal machine learning system will be evaluated on the same domain data used for training to maximize performance, however there exist domains for which collection of source test data is prohibited by cost, ethical restrictions or policy.  This introduces a domain shift between training and testing that will affect performance, and must be mitigated by fine-tuning on source domain data or domain adaptation.  

Visual AI has the potential to give us helpful and personalized insights into the rich patterns of our daily life.  However, we humans spend the majority of our time in what may be called the {\em dark domains} of visual AI.  These are locations or data sources that are not broadly exposed to visual AI today due to privacy concerns, robustness issues or a lack of datasets.  For example, third person viewpoints of people in private or shared spaces could enable helpful applications of ambient healthcare \cite{Haque20}\cite{Martin20},  wellness monitoring \cite{Berridge20}\cite{pmlr-v68-haque17a} or ethical security \cite{Corona2021WACV}\cite{Rai21}\cite{ji2020action}\cite{Byrne2020bmvc}.   
However, these spaces contain our private data, and privacy regulations (e.g. GDPR, BIPA, CPRA) require that visual AI address collection and protection of private data by design.

Consider the collection of long duration videos from a third-person viewpoint in private spaces.  These videos may require hundreds or thousands of hours per camera to capture rarely occurring activities, all while watching and recording the intimate details of the private lives of subjects.  This introduces (i) an engineering challenge of sharing this huge volume of data, (ii) a sparsity challenge of efficiently sharing data that is mostly empty video between interesting activities, (iii) a long-tailed challenge of collecting data of rare activities that may never occur organically, (iv) a bias challenge where people who have consented to recording and know they are being recorded change their behavior \cite{Berridge20} and (v) an ethical challenge of whether we should share data from private spaces in the first place.

\begin{figure*}
\centering
\includegraphics[width=\figwidth]{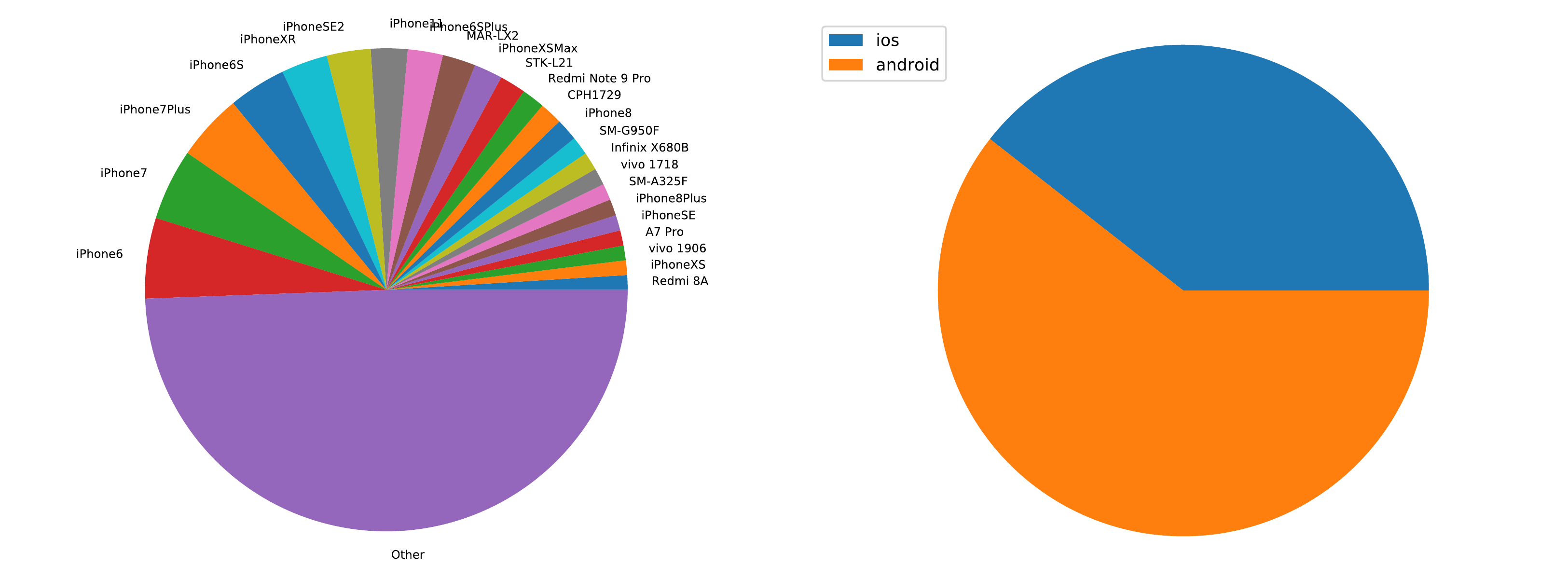}
\caption{CAP mobile devices.  Distribution of mobile devices and mobile OS from collector submissions.  This shows that there is a wide variety of mobile devices used to collect the CAP dataset, with a slight preference for Android devices.}
\label{f:cap_mobile_devices}
\end{figure*}

This suggests that direct collection and distribution of in-domain data for these dark domains should be avoided. There exists a tradeoff between the volume of data curated vs. the quality of the data collected for target domain deployment.  Our fundamental hypothesis is that collecting a large amount of annotated data in a closely related source domain (e.g. short duration, on-demand, third-person videos), then deploying the trained system to an adjacent target domain with privacy restrictions (e.g. long duration third-person videos in private spaces) will enable ethical deployment of a trained system for dark domains.  However, a key challenge is collecting domain adjacent data for these dark domains without compromising visual AI system performance.  This issue will be explored in section \ref{a:benchmark}.


\subsection{Dataset Collection}
\label{s:cap_collection}

Dataset collection is the process of defining, recording, annotating, verifying and distributing a dataset that achieves the design goals in section \ref{a:design_objectives} while addressing the key challenges in section \ref{a:cap_design_challenges}.  The dataset collection leverages the Collector platform introduced in section \ref{s:collector} to collect the data.  In this section, we address design issues related to setting up the Collector campaign.

\medskip
\noindent {\bf Collection Campaign}.  The collection campaign was specified on the Collector platform as follows:  

\medskip
\begin{squishlist}
\item {\bf Collections}.  The campaign specification includes 842 unique collection types, each specifies one of 512 activity labels with person interaction or interactions with 157 object types.  288 of 842 collection types were specified to be collected so that the subject is ``facing away'' from the camera, so that the back of the subject is more visible than the front, such that the activity may be partially occluded to increase diversity.
\item {\bf Repetitions}. Collectors were tasked with repeating activities between 5-10 times in each submission.  This provides in-video data augmentation, but rather than synthetic augmentations (e.g. mirror, crop, scale).  We ask collectors to perform the activity slightly differently each time, enabling natural data augmentation.
\item {\bf Third-person viewpoint}. Collectors were tasked with collecting all videos from a third-person viewpoint, looking slightly down on the scene from above, to model a security camera on the wall or ceiling.  
\item {\bf Physically stabilized}. 38 of 842 collection types were specified to be physically stabilized.  We tasked collectors to rigidly mount their device to a simple triangular cardboard stand and put it on a flat surface up high looking down (e.g. a shelf, a stack of boxes, a ladder).  Collectors record their subject (or themselves) performing the scenario, then they use the in-app editor to annotate when and where the activities were performed.  
\item {\bf Temporal activity detection}. 87 of 842 collections were specified to be collected to support temporal activity detection. We instruct the collectors to choose from a list of 11 activities to perform in a natural sequence called a ``scenario''.  For example, one scenario is ``Come inside from the cold'' which includes the activities of entering a room and taking off winter outerwear.  The subject performs at most 11 activities in any order they choose, then the collector records the scenario, and edits in-app to annotate the performed activities.  
\end{squishlist}
\medskip

\begin{figure}
\includegraphics[width=\fighalfwidth]{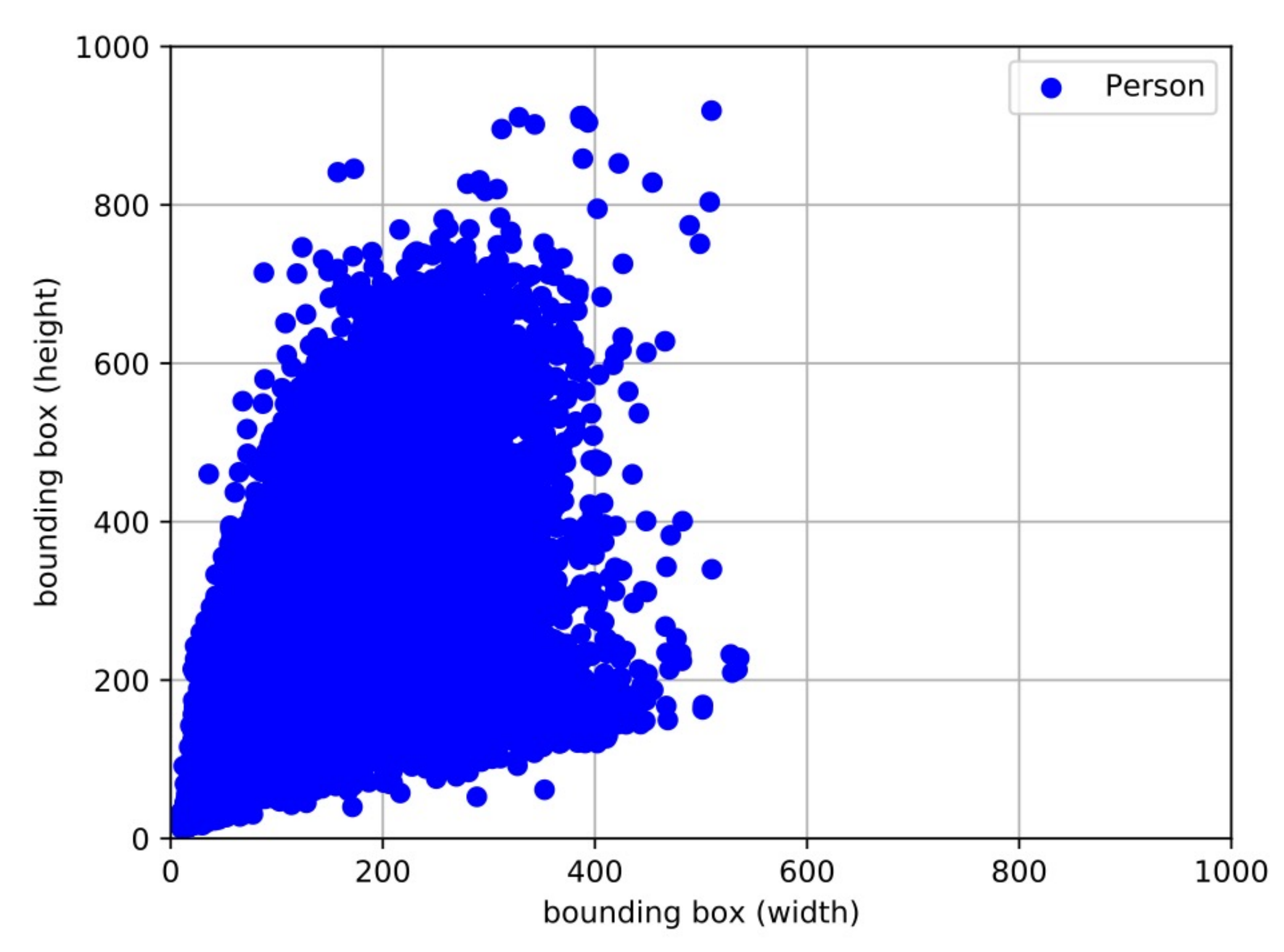}
\caption{CAP 2D Bounding Box Distribution showing the size variation of labeled people boxes.}
\label{f:cap_bounding_box}
\end{figure}

The overall collection statistics are shown in figure \ref{f:cap_statistics} which shows the submission locations of collectors worldwide, along with the label histogram in figure \ref{f:cap_histogram}.  Figure \ref{f:cap_details_1} describes additional supporting details about the dataset including collector submission frequency, bounding box size distribution (Figure \ref{f:cap_bounding_box}), mean duration per collection type  and activity density per video, and mobile OS and device type distribution (Figure \ref{f:cap_mobile_devices}).

\begin{figure*}[t!]
\centering
\includegraphics[width=\figwidth]{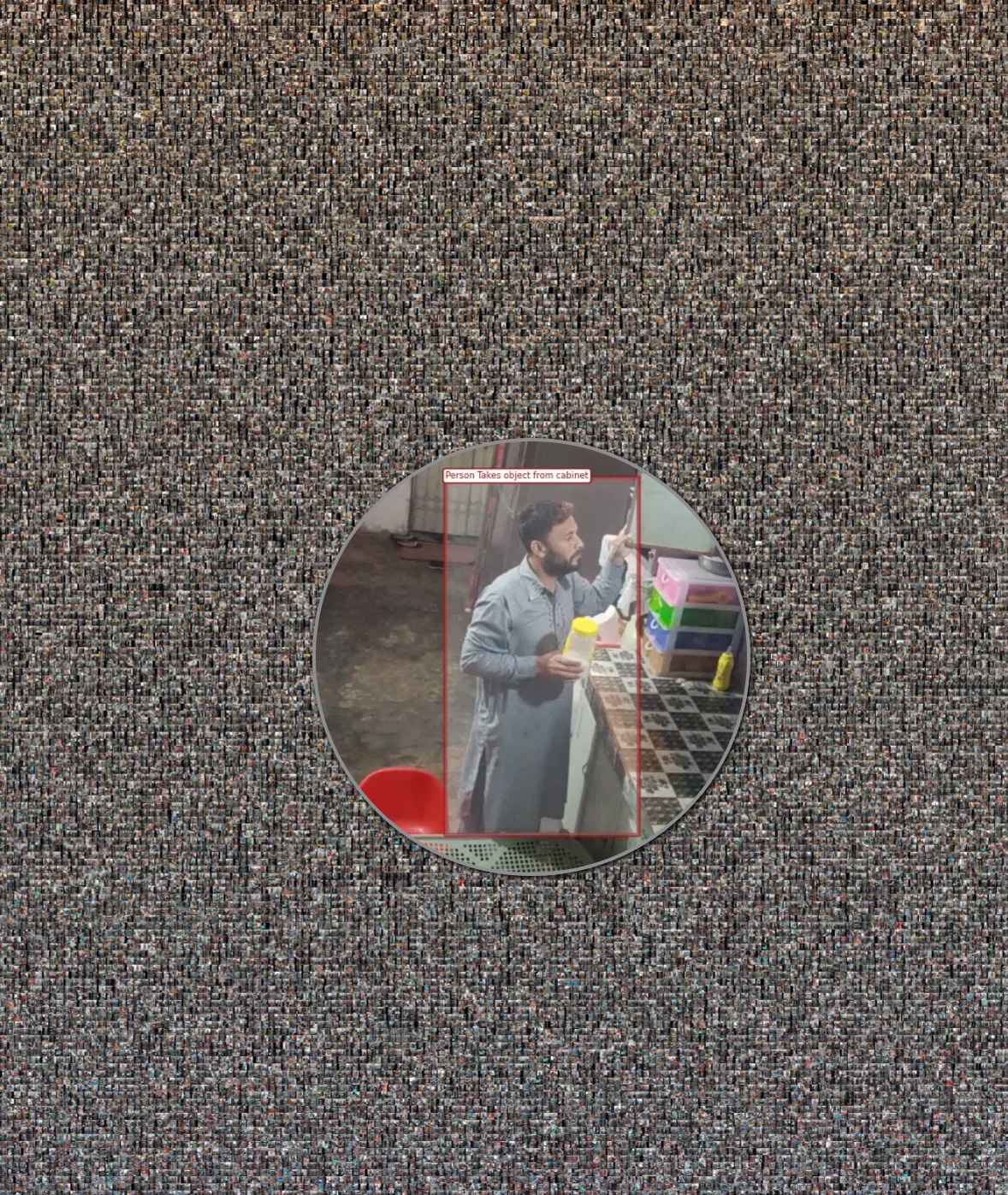}
\caption{CAP Dataset Explorer.  This visualization shows a 1\% sample of the CAP dataset, tightly cropped spatially around the actor and cropped temporally around the fine-grained activity being performed. The full dataset includes the larger spatiotemporal context in each video around the activity, and the complete set of activity labels. This 
\iftoggle{doubleblind}{open source visualization tool}{\href{https://visym.github.io/cap}{open source visualization tool}} 
can hover over a specific video in the montage to show a high resolution animation.  The explorer can be sorted by category or color, and shown in full screen.}
\label{f:cap_explorer_large}
\end{figure*}

\begin{figure*}[t!]
\centering
\includegraphics[width=\figwidth]{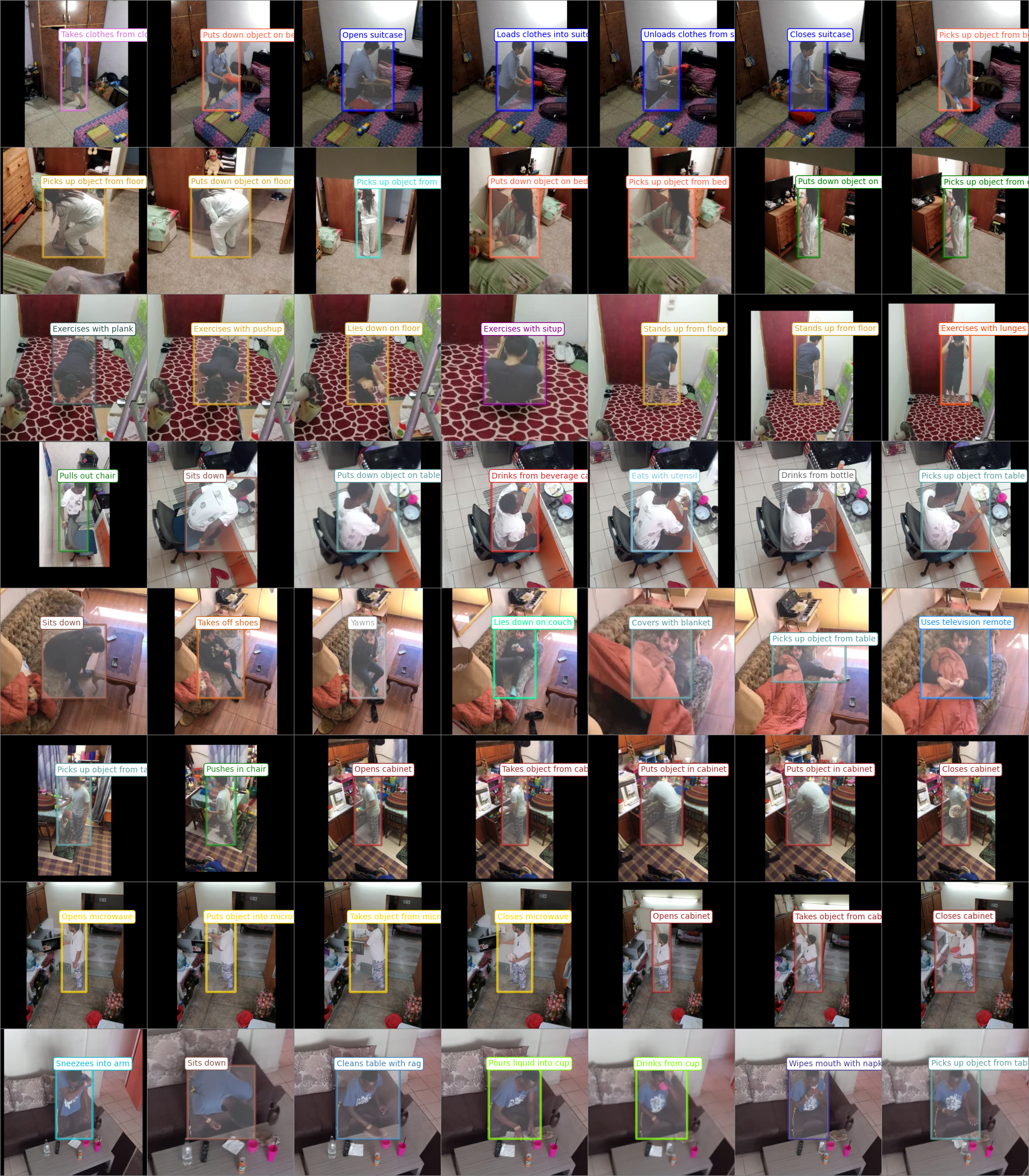}
\caption{Activity detection benchmark examples.  This visualization shows eight videos sampled from the activity detection task.  Each video is collected as a ``scenario'', which is a sequence of seven to eleven activities performed in any order that the subject in the video chooses.  Each row shows the middle frame of the temporal activity annotation, spatially cropped around the primary actor and annotated with an activity caption. The scenarios by row are: ``Get ready to go on a trip'', ``Organize a cluttered room by putting things away where they belong'', ``Exercise'', ``Eat a snack'', ``Sit and watch TV'', ``Set the dinner table'', ``Cook a meal in the microwave'' and ``Drink a glass of water''.  For example, the scenario ``Get ready to go on a trip'' includes the activities: take clothes from a closet, put objects onto the bed, open suitcase, load and unload clothes into a suitcase, close a suitcase.  A system evaluated on the activity detection task is required to temporally localize these activities in untrimmed clips.}
\label{f:cap_ad_montage}
\end{figure*}

Finally, a visualization of the scale of this dataset is shown in figure \ref{f:cap_explorer_large}, which shows a montage of less than 1\% of the videos, along with a visualization tool to interactively explore the dataset.  
Figure \ref{f:cap_ad_montage} shows a montage of ground truth examples of the activity detection task.  This shows samples of frames from the activity detection task that shows the sequences of activities that a collector is tasked to perform in eight scenarios. 

\begin{figure*}[t!]
\centering
\includegraphics[width=\figwidth]{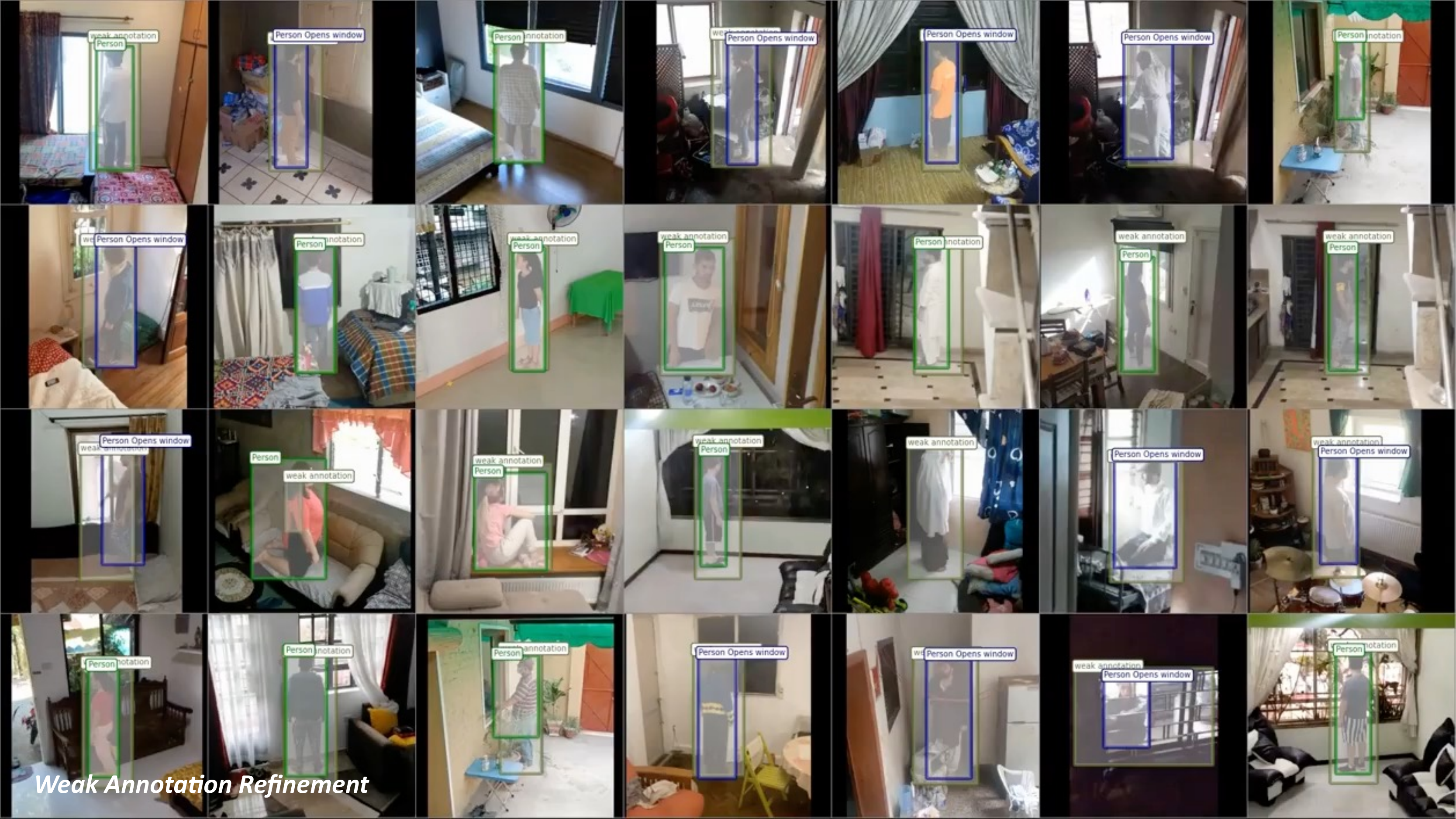}
\caption{Dataset post-processing by 
\iftoggle{doubleblind} {weak annotation refinement}{\href{https://youtu.be/Qyth4z3XlYk}{weak annotation refinement}}.
The human annotation (captioned ``weak annotation'') shows the box defined by the collector in-app while recording, which is generally centered on the subject, but may not be tight around the limbs.  The refinement (captioned ``Person'') selects an person track using object detection proposals that optimally overlaps with the weak annotation.  Both annotations are available in the public dataset release.}
\label{f:collector_post_processing_refinement}
\end{figure*}

\subsection{Dataset Format}
\label{a:cap_format}

The CAP dataset is publicly available for download.  In this section, we provide details on the dataset format, including the selection of the label naming format, post processing for bounding box improvement and background stabilization and additional video metadata.

\medskip
\noindent {\bf Activity Label Nomenclature}.
The label format for atomic activity labels follows the following compositional structure: ``noun verb adjective noun''.  For example, {\em person opens car door} or {\em person opens refrigerator door}.  This structure provides a consistent naming structure for atomic activities, that allows unambiguous description of an atomic activity with optional adjective extensions to provide specificity for person-object or person-person interactions.  Furthermore, we specify the hierarchical label structure using only the ``noun verb'' components of this label, with the remaining label components specifying the within-class or fine-grained variation.  

The label format for this dataset is in contrast with other large scale activity datasets.  For example, the caption label style of Charades \cite{sigurdsson2016hollywood} or Something-Something v2 \cite{1706.04261} describes a label as a phrase or short natural language narrative of the contents of the video (e.g. Putting a book somewhere, Approaching something with your camera).  The verb only label style of Moments in Time \cite{Monfort2020}, AVA \cite{gu2018ava}, HMDB \cite{Kuehne2011}, Kinetics \cite{kay2017kinetics} and ActivityNet \cite{caba2015activitynet} describes an activity in terms of the verb being performed, largely independent of the object being interacted with or actor performing the verb.  Finally, our nomenclature goal is to provide a more intuitive label description than alternatives previously deployed, such as wordnet synsets \cite{imagenet_cvpr09}.
Our goal is to provide more specific representation of the within class variation of activity classes, by exploring the actor, person-object and person-person interactions as represented in the label name.  This label nomenclature is closely related to the Multiview Extended Video with Activities (MEVA) class naming \cite{Corona2021WACV}.

The label format for this dataset is also in contrast with {\em open vocabulary datasets} \cite{Ego4D2021}\cite{gupta2019lvis}.  In this style of dataset collection, raw data is recorded in a target domain explicitly without a target task in mind for this data.  The raw data is collected, then it is post processed by an annotation team to provide labels or natural language captions for the data that was collected for a task defined after collection was performed.  For example, the Ego4D dataset \cite{Ego4D2021} recorded egocentric video from first person perspective of wearers going about their daily lives, which is captioned after collection.  This data provides a sample of the common activities that were performed during the recording period, but this does not provide the training data needed for supervised learning.  Similarly, the LVIS dataset \cite{gupta2019lvis} for large vocabulary long tailed object recognition collects cluttered images in natural settings and asks annotators to achieve consensus for labeling all of the objects that are present in these images.  This can achieve dense labels in naturally occurring images, however it cannot achieve balanced datasets.

\begin{figure*}[t!]
\centering
\includegraphics[width=\figwidth]{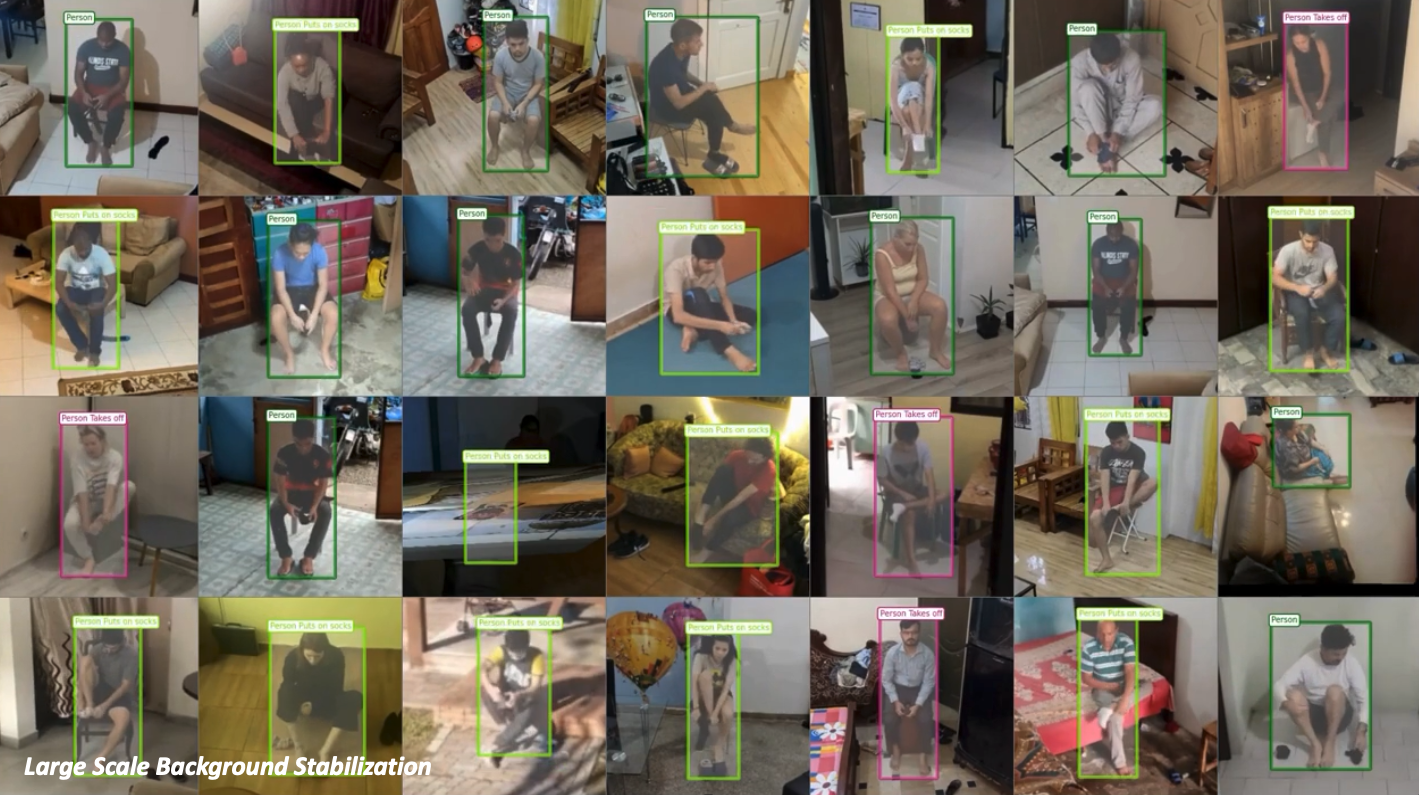}
\caption{Dataset post-processing by 
\iftoggle{doubleblind} {background stabilization}{\href{https://youtu.be/Iyo4fRLR65Q}{background stabilization}}.
We use a flow based affine stabilization method to align each frame to the first frame of the video to enforce that that the background is unmoving, as if the video was collected from a rigidly mounted camera.  The background stabilization is best shown in video, comparing 
\iftoggle{doubleblind} {unstabilized handheld video}{\href{https://youtu.be/Je91vWjSHpo}{unstabilized handheld video}}
collected from mobile devices to 
\iftoggle{doubleblind} {background stabilized video}{\href{https://youtu.be/Iyo4fRLR65Q}{background stabilized video}}.  
Bounding boxes are affine transformed using the affine stabilization transformation to align boxes to the stabilized video.}
\label{f:collector_post_processing_stabilization}
\end{figure*}

\medskip
\noindent {\bf Weak Annotation}.
Videos are post-processed after collection to include an optional step of weakly annotated bounding box refinement.  Weak annotation refinement is the process improving the bounding box provided in-app by the collector.  The annotation by the collector provides a weak label that is collected quickly and easily, which coarsely overlaps the true object, with minor misalignment errors.  Then, a pretrained object detector selects an optimal low confidence proposal that maximizes overlap and confidence with the weak annotation.
This strategy significantly reduces the cost of large scale annotation by performing the annotation while recording videos, with error correction in post-processing. 

The weak annotation refinement is performed as follows.  Videos are processed with a low confidence object detector for the target actor in the video (e.g. a person detector) forming low confidence object proposals.  Note that this strategy requires a pretrained object detector for the target class.  Next, proposals per frame are grouped using maximum intersection over union (IoU) assignment forming object tracks for each object instance in the video.  Next, given a sequence of object bounding boxes from a human annotator, object tracks are rescored to compute the framewise product of proposal confidence and IoU with the annotated box, followed by a mean score over all frames in the track.  Finally, the object tracks are sorted by this rescored confidence, and the track with the highest score selected as the weak annotation refinement.  This selects the object track that maximally overlaps the weak annotation with highest confidence, forming a weak annotation refinement.  This procedure assumes that the annotated box from the collector overlaps the primary actor by at least 50\%, which is enforced during the human review process.  Both the refined box and the collected box are exported in the dataset release.  The weak annotation code is 
\iftoggle{doubleblind} {open source}{\href{https://visym.github.io/vipy}{open source}}.

Figure \ref{f:collector_post_processing_refinement}  shows an example of the weak annotation refinement.  In this montage, each montage element is a frame from a collected video.  We show the bounding box annotated by the collector live during collection with the caption ``weak annotation''.  This provides the weak label which is improved into the refined bounding box with the caption ``Person'' (or ``Person Opens Window'' if the subject was in the process of performing this activity in this video).  This shows that the refinement procedure creates boxes that are tight around the torso and limbs, even when the subject is in a atypical pose or partially occluded in the scene.  However, this derived method can still introduce rare assignment errors (e.g. row 3 column 2), or missing annotations if there is no overlap between tracks and annotations, so the end-user should be aware of possible corner cases and fall back on the weak annotations as needed.  The weak annotated refinement is best shown in a video, showing 
\iftoggle{doubleblind} {weakly annotated bounding boxes}{\href{https://youtu.be/Qyth4z3XlYk}{weakly annotated bounding boxes}}
sampled from the CAP dataset.

\medskip
\noindent {\bf Background Stabilization}.
Background stabilization is the process of stabilizing the camera to the first frame so that the background is unmoving.  This strategy reduces the cost of large scale dataset collection by not requiring that the cameras be rigidly mounted on a tripod, since few freelancers have tripods with mobile device mounts available for use.  The collector can record videos handheld, which are post-processed to stabilize the videos as if they were rigidly mounted and unmoving.  

The approach for background stabilization is affine stabilization to frame zero using multi-scale optical flow correspondence with foreground object keepouts.  This pipeline supports optical flow based stabilization of video which reduces the artifacts due to hand-held cameras to stabilize the background. Remaining artifacts are due to non-planar scenes, rolling shutter distortion and subpixel optical flow correspondence errors. The stabilization is only valid within the tracked actor bounding box for small camera motions. Large motions will introduce stabilization artifacts due to non-planar scene effects and should be filtered prior to usage. The stabilization artifacts will manifest as a slightly shifting background relative to the actor which may affect flow based methods.  Finally, the approach transforms all bounding boxes to be aligned with the stabilized video, and includes the stabilization residual in video metadata to enable filtering stabilization with poor alignment.  The stabilization code is
\iftoggle{doubleblind} {open source}{\href{https://visym.github.io/vipy}{open source}}.

Figure \ref{f:collector_post_processing_stabilization} shows an example of the background stabilization.  The background stabilization is best shown in video, comparing 
\iftoggle{doubleblind} {unstabilized handheld video}{\href{https://youtu.be/Je91vWjSHpo}{unstabilized handheld video}}
collected from mobile devices to 5Hz 
\iftoggle{doubleblind} {background stabilized video}{\href{https://youtu.be/Iyo4fRLR65Q}{background stabilized video}}.  
Observe that the background stabilized video in these YouTube links has the background unmoving as if the video was collected from a rigidly mounted static camera.

\medskip
\noindent {\bf Privacy and Consent}.
We require that all subjects review a consent form, and provide their informed consent for the dataset collection.  This consent form is Institutional Review Board (IRB) approved and describes how the data will be collected, what data will be collected, how it will be shared and who it will be shared with.  The all subjects consent to their likeness to be shared in publication material and a publicly accessible dataset release for the purposes of visual AI research.  Each subject is required to provide a video consent, which is a selfie video of this subject stating that they consent to the video collection.  This selfie video is used by the review team to compare that the subject recorded in the video is the consented subject.  This enables rejection of fraudulent video submissions of non-consented subjects.  An example of this selfie consent video is shown in the upper left of the review interface in figure \ref{f:collector_review_interface}.  IRB consent forms shown in-app are customizable for local IRBs.

Non-consented subjects in the video field of view have their faces blurred out in-app prior to submission.
All subjects with visible personally identifiable information (PII) in the videos have consented to having their PII shared for the purposes of visual AI research.  Non-consented subjects are those subjects that are not within the foreground box of the primary actor.  All non-consented subjects have their faces blurred out on-device in the mobile app prior to submission.  This feature is currently enabled only in the iOS release.

\medskip
\noindent {\bf Collection Metadata}.
The collection platform includes the following additional metadata released for each video:

\smallskip
\begin{squishlist}
\item {\bf Program name/ID.} This is the name of the campaign under collection.  For the CAP dataset, the program name was either PIP or MEVA.  Program names (and associated program ID) are globally unique on the collector platform.
\item {\bf Collection name/ID.} This is the description of the collection that is given to the collectors.  Collection names are globally unique to a program.  
The ID is a globally unique string that uniquely identifies this collection.  
\item {\bf Collection date.} This is the timestamp when the video was collected in the local timezone of the device.
\item {\bf Geolocation.} This is the region of the world that the collection was recorded (with consent).  The geolocation is captured with the field ``ipAddress'' which is the public IP address of the internet service provider of the device.  IP geolocation services can be used to convert this IP address to a geographic region.
\item {\bf Collector ID.} A globally unique identifier for the registered collector who was logged into the mobile app and is recording the video.  Collector IDs have been anonymized to avoid association with email addresses.
\item {\bf Subject ID.} A globally unique identifier for the consented subject who is in the video.  This is the ID of the subject who was identified at consent time.
\item {\bf Mobile device.} The mobile OS (Android, iOS) and device hardware that was used to collect the source video.  
\item {\bf Frame rate.} The frame rate in fractional frames per second at which the original video was collected.
\item {\bf Video dimensions.} The video (height, width) in pixels.  
\item {\bf Rotation.} The rotation state (e.g. landscape, portrait, left, right) of the device when recording.
\item {\bf Blurred faces.} The number of non-consented faces blurred on-device prior to video submission.  This feature is enabled for iOS devices only.  Non-consented faces are those not within the collector annotation box.
\item {\bf App version.}  A version string that identifies the release version of the mobile app used to collect this video.  
\item {\bf Video ID.}  A globally unique ID that uniquely identifies this video on the collector platform.

\end{squishlist}
\smallskip

\medskip
\noindent {\bf Bias Engineering}.  Our collections organize activities into groups to introduce diversity in the scene. For example, we specify to the collectors to load and unload both from a trunk and from a rear door of a vehicle to help introduce within-class diversity. Also, we introduce joint activities such as "Leave this scene while talking on a phone".  Finally, we specify that collectors should be facing or facing away from the camera to introduce more pose diversity.  The full list of collection names are self explanatory and may be filtered to remove variants that may not reflect the target domain bias, or which do not satisfy the assumptions of the loss function of the target system.

\medskip
\noindent {\bf Temporal Padding}.  All data is distributed in a clipped or padded form.  The clipped dataset includes activities that are tightly temporally cropped around each activity, such that the duration of the clip is the duration of the video.  The padded dataset temporally pads each video to $\ge3$s, along with the metadata to recover the tight clip.  This provides additional video context for each training video to support the temporal assumptions of a target system or additional data augmentation.

\medskip
\noindent {\bf Recommended Splits}.
The training, validation and test set splits are generated using an 80/10/10 split strategy controlling for collector ID.  The dominant source of bias in the dataset is the effect of collectors submitting different activities in closely related locations, such as the same house or yard.   We seek to avoid the same collector providing videos for both training and testing, in order to create a more realistic testing scenario.  To achieve this, we randomly split the collector IDs into 80/10/10 splits, then assign all videos submitted by this collector into the corresponding training/validation/test sets.  The test set is sequestered for leaderboard evaluation purposes, and videos are available for download behind a license agreement restricting redistribution.  

Figure \ref{f:cap_details_1} shows the distribution of videos per collector which shows a falloff in submission frequency and allows a random sample of collectors to avoid unbalanced video distribution in the dataset. The final split enforces that collector IDs are disjoint between training, validation and test.  The splits are included in the release metadata. 

\medskip
\noindent {\bf Annotation format}.  Dataset management includes processing annotations and pixels for a large number of videos.  In order to streamline this data processing pipeline, we have developed the open source 
\iftoggle{doubleblind} {\href{https://}{REDACTED package}}{\href{https://visym.github.io/vipy}{vipy package}}.
\iftoggle{doubleblind}{{\bf REDACTED}}{Vipy} 
is a Python package for representation, transformation and visualization of annotated videos and images.  
\iftoggle{doubleblind}{{\bf REDACTED}}{Vipy} 
provides tools to apply transformations such as downsampling, padding, scaling, cropping and rotating so that the annotations are transformed along with the pixels.  The 
\iftoggle{doubleblind}{{\bf REDACTED}}{vipy} 
annotation format is open JSON designed for representation of activity and object annotations in video.

\subsection{Benchmark Research Questions}
\label{a:research_questions}

This dataset provides the data to answer the following research questions.  The answers to these research questions are provided in section \ref{s:cap_benchmark}.

\medskip
\noindent {\bf Fine grained categories.}  Is there an improvement when training using fine grained categories (e.g. {\em person picks up object from floor} vs. {\em person picks up object from table}) vs. coarse grained categories (e.g. {\em person picks up object}) when testing coarse grained activity classification and temporal activity detection?  

\medskip
\noindent {\bf Collection diversity.}  Is there an improvement when training with explicitly engineered within-class diversity (e.g. two biases are are controlling for are activity instances collected with actor pose explicitly "facing away" from the camera, and instances collected in a crowded scene with nearby occluding people).  

\medskip
\noindent {\bf Collector diversity.}  What is the relationship between the number of unique collectors in the training set vs. test set performance?   How many unique collectors do we need?  Does it help to have one collector doing many collections in the same location and clothes?

\medskip
\noindent {\bf Stabilization.}  What is the effect when training software background stabilization from handheld collection on rigidly mounted videos?  Can we correct for this domain shift through software stabilization?

\medskip
\noindent {\bf Video data augmentation.}  Is there a benefit using actor data augmentation (e.g. collectors repeating activities slightly differently each time) vs. synthetic data augmentation (e.g. crops, scales, rotations)?

\subsection{Benchmark Evaluation}
\label{a:benchmark}

Performance benchmarking is the specification of an evaluation methodology, task and dataset along with a baseline system design to evaluate system performance.  However, section \ref{a:cap_design_challenges} discussed that this is impractical for long duration third person videos.    
For example,  we may collect thousands of hours of video from a security camera without ever collecting an organic instance of {\em person puts on shoes}.  How do we realistically benchmark a task where the labels to evaluate may never occur?  Furthermore, even if we did have enough instances of all the target labels, how do we address the ethical concerns of publicly sharing long duration video of the private daily lives of humans without this being interpreted as a fishbowl or worse, exploiting people to create a human zoo?

Section \ref{a:cap_design_challenges} addressed this key challenge by introducing {\em domain adjacent benchmarking}.  In this strategy, we collect test sets that are from the required viewpoint, but with actors performing the test activities in short bursts, rather than real subjects going about their daily lives.  This provides performance evaluation of the dark domain (e.g. third person, long duration videos collected in private spaces) in a closely related adjacent domain (e.g. third person, short duration videos acted in shared spaces).  The test data in the adjacent domain can be collected and distributed ethically, and performance evaluation on the domain adjacent data is used as a surrogate for the dark domain.

However, this strategy exposes a fundamental tradeoff between bias and diversity.   As discussed in section \ref{a:cap_format}, the Collector platform controls diversity through {\em bias engineering}, where we explicitly request collection parameters to encourage diversity of videos submitted.  This allows collection of rare activities that may not occur organically, however these subjects know they are being recorded.  Their payment depends on their submission passing verification, which can encourage movements in an exaggerated or unnatural manner.  This introduces an ``acting bias'' into the dataset, that would not be present in the target dark domain.  More generally, we have observed the following biases when collecting domain adjacent test data:

\smallskip
\begin{squishlist}
\item {\bf Duration bias.}  All videos are limited to a maximum duration of 45 seconds.  This duration bound is due to the practical limitations of uploading large videos from cellular data connections in less developed countries, and the design goal of avoid long duration videos where nothing interesting occurs.  
\item {\bf Actor bias.}  People sometimes perform in an exaggerated manner when they know they are being filmed, resulting in awkward or too well-framed scenes.
\item {\bf Handheld bias}. 95\% of the dataset is collected handheld which leads to moving camera artifacts.  Approximately 5\% is collected physically stabilized, and all videos are background stabilized in post-processing. 
\item {\bf Sequence bias}. Activities are often performed in a proscribed sequence.  It is difficult for a subject to remember all the things they are supposed to do, so they often perform the same activities in the same order.
\item {\bf Center bias}. Actors are always in the video center and never occluded by image boundary.  This is due to the in-app annotation methodology where the bounding box is specified by keeping the subject in the box in the center of the camera while recording.
\item {\bf Consent bias}. All our subjects are required to consent to using their personally identifiable information for improving visual AI research.  As such, we do not have videos of people in large crowds due to the need to get consent from every subject.
\end{squishlist}
\smallskip

Finally, validation of the domain adjacent benchmarking strategy requires an evaluation of the same system on a private (and unreleasable) test set from the dark domain as compared to the public test set in the adjacent domain.  This validation is important to establish trust that performance on the domain adjacent test set is predictive of performance in the dark domain, and an initial study was performed in section \ref{s:actev} on a subset of the CAP labels.  However, additional work is needed to characterize the performance of the full fine-grained label set on security video.

\begin{figure*}[t!]
\centering
\includegraphics[width=\figwidth]{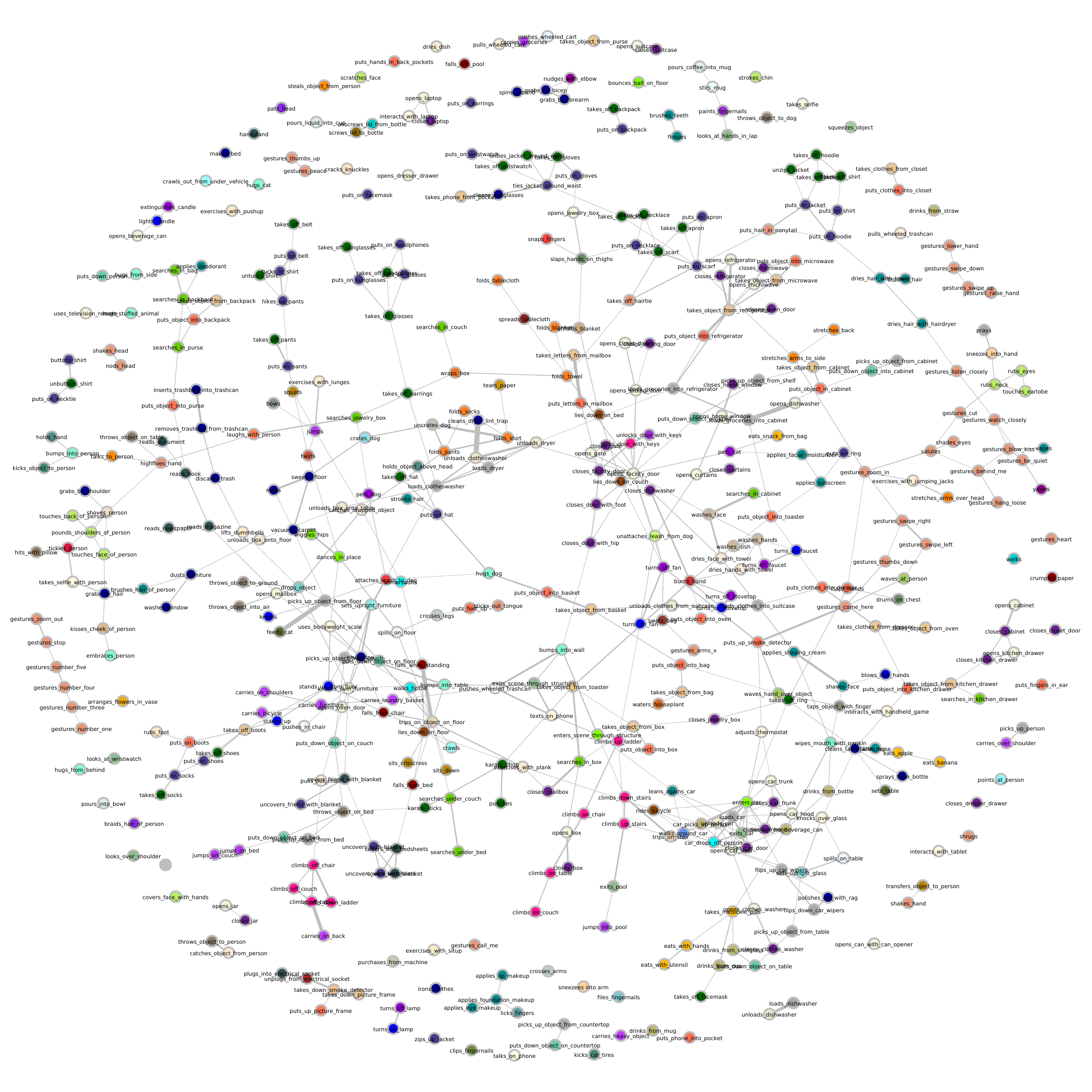}
\caption{A confusion graph embedding for the activity classification task in the CAP Benchmark showing edges connecting commonly confused fine-grained activity labels.  Node colors correspond to coarse-grained activity labels, and edge thickness corresponds to the confusion weight.  The graph embedding was constructed by computing a confusion matrix on the activity classification test set, such that each row of the confusion matrix was normalized to sum to one, then thresholded at 0.07 removing self edges.  This forms a sparse adjacency matrix of pairs of labels that are commonly confused.  This {\em confusion graph} visualization provides a 2-d graph embedding of neighborhood structures for commonly confused labels, such that the 2-d graph layout was constructed using a force-directed graph embedding to maintain constant edge length and minimize edge crossings.
This provides a visualization of the visual similarity of fine-grained activity classes as compared to the semantic similarity as specified in the coarse-grained label space.  We recommend zooming into the PDF to see the node labels.}
\label{f:cap_benchmark_confusion_graph_large}
\end{figure*}

\begin{figure*}
\centering
\includegraphics[width=\figwidth]{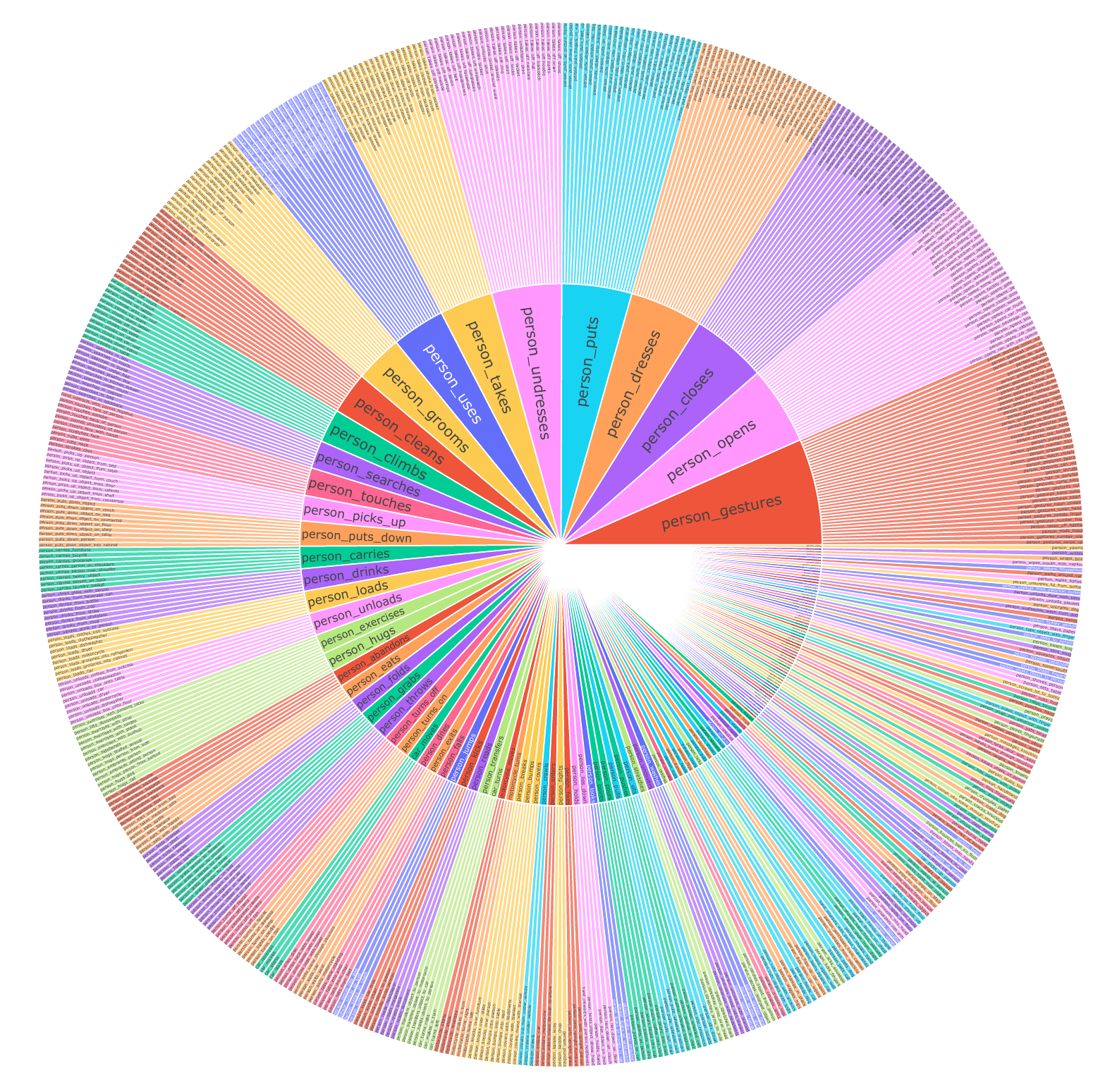}
\caption{CAP hierarchical label structure, visualized as a circular tree with outer fine labels grouped by inner coarse labels. We recommend zooming into the PDF to view the labels.}
\label{f:cap_labels}
\end{figure*}

\end{document}